# *Cognition* is All You Need
## *The Next Layer of AI Above Large Language Models*

Nova Spivack[1], Sam Douglas[1], Michelle Crames[1], Tim Connors[1]

[1] Mindcorp, Inc (www.mindcorp.ai)
contact@mindcorp.ai
www.mindcorp.ai
www.linkedin.com/company/mindcorp-ai
twitter.com/mindcorpai

# Contents

# Abstract

Recent studies of the applications of conversational AI tools, such as chatbots powered by large language models (LLMs), to complex real-world knowledge work have shown limitations related to reasoning and multi-step problem solving. Specifically, while existing chatbots simulate shallow reasoning and understanding they are prone to errors as problem complexity increases. The failure of these systems to address complex knowledge work is due to the fact that they do not perform any actual cognition. In this position paper, we present a higher-level framework ("Cognitive AI") for implementing programmatically defined neuro-symbolic cognition above and outside of large language models. Specifically, we propose a dual-layer functional architecture for Cognitive AI  that serves as a roadmap for AI systems that can perform complex multi-step knowledge work. We propose that Cognitive AI is a necessary precursor for the evolution of higher forms of AI, such as AGI, and specifically claim that AGI cannot be achieved by probabilistic approaches on their own. We conclude with a discussion of the implications for large language models, adoption cycles in AI, and commercial Cognitive AI development.

# Introduction

As the landscape of artificial intelligence continues to evolve towards increasing levels of intelligence, a new architectural paradigm is emerging: **Cognitive AI ("CogAI").** In this position paper we will explore the distinctions between Conversational AI and Cognitive AI, with a focus on the key functional architecture components and requirements for Cognitive AI systems that are capable of doing complex knowledge work.

*Cognitive AI* represents a foundational shift in how AI systems are conceived, developed, and deployed. It is a distinct approach - focused around a new neuro-symbolic reasoning layer that works above Large Language Models (LLMs).

Transformer-based LLMs (or similar probabilistic language models) will never be able to replicate what Cognitive AI is capable of, and in fairness, they were not designed to. However, this fact is not well-understood and has led to the widespread misconception that innovation on the model level will continue to yield major advances. LLMs will never actually get us to a significantly more advanced level of AI, because of their many inescapable limitations.

LLMs are an essential enabling technology for  Cognitive AI, and indeed, without them it cannot emerge or function. But the question is when will LLMs be "good enough" for Cognitive AI? Our answer is that they are already good enough. The current generation of large foundation models, plus growing diversity of more specialized open-source models, is already sufficient to meet the intelligence needs of the Cognitive Layer. Further improvements to LLMs, or any other Conversational AI level technologies, will only yield limited benefits.. Advancements in Cognitive AI will be more profound and will have more impact.

Key to the Cognitive AI paradigm is the representation and implementation of cognitive processes in a new *"Cognitive Layer"* that sits above the Conversational Layer where LLMs reside. The Cognitive Layer introduces a range of cognitive functions and capabilities which are beyond the reach of LLMs, yet use LLMs as tools.

While the Cognitive Layer utilizes LLMs extensively, it is a higher-order layer of intelligence above the intelligence inherent in LLMs, giving it meta-level capabilities that far exceed what LLMs can do on their own. By utilizing the Cognitive Layer, Cognitive AI architectures are able to implement the higher levels of cognition that are necessary for real-work knowledge work, which in-turn is a precondition for mainstream adoption of AI.

This phase transition is not merely an incremental improvement but a rethinking of AI's approach to performing complex cognitive tasks, combined with a new architectural paradigm, that together push the envelope of what machines can understand, and how they can interact with the world around them.

Cognitive AI offers a new frontier for research, development, IP and commercial applications that will be larger than the conversational AI frontier.

*By 2030, if not sooner, we predict it will disrupt the AI landscape by shifting the focus of innovation and competition to a new playing field.*

Cognitive AI provides a practical way to utilize the many prior decades of research and development in AI that preceded Conversational AI, above the Conversational Layer.

In addition, Cognitive AI is social. Human intelligence does not happen in a vacuum, it is a social process. Learning is a social process, as are nearly all human activities. It follows that the majority of human cognition is social, and the same goes for knowledge work.

It is necessary for any system capable of doing high-level knowledge to be built for cognition across social relationships. Specifically Cognitive AI is actually a form of collective cognition that leverages relationships among networks of agents - whether they be humans or software agents - to think, solve problems, innovate, and do knowledge work together.

Collective cognition requires that all cognitive processes be at least potentially social and collaborative, if necessary. Whether it is storing and retrieving memories or expertise across relationships, or teaming up to solve a specific problem, Cognitive AI systems have to be able to leverage both distributed networks of human and machine intelligence. To do this effectively means these capabilities should not be "bolt-on" afterthoughts but rather they should be intrinsic to how such systems work.

The combination of both machine and human intelligence enable a higher level of cognition that goes beyond what AI can ever produce by itself. We call this *"exponential intelligence."*

Exponential intelligence is defined as a higher form of intelligence that emerges when human and machine intelligence are combined such that increasingly large and complex many-to-many cognitive processes can take place.

By enabling a deeper symbiosis (exponential intelligence) between human and machine intelligence, Cognitive AI will radically advance how people do knowledge work. Here we can view Cognitive AI as a partner with, not a replacement for, human knowledge workers.

Cognitive AI will enable humans to become more productive at knowledge work, and also to become better at it. In particular, Cognitive AI will make it possible for larger and more complex knowledge work to be completed by fewer people.

This will not only advance knowledge worker capabilities but it will also enable them to work on classes of problems that were previously thought to be too complex or difficult to do at all. In other words, Cognitive AI will move the frontier, bringing previously unattainable levels of cognition within reach of individual knowledge workers. This can help humanity solve the complex problems we face in the future.

Without adopting Cognitive AI, the field of AI can never achieve the level of reasoning required for complex knowledge work (Thórisson, 2020, Thórisson & Talbot, 2018). This means that attempts to use LLMs on their own to achieve artificial general intelligence ("AGI") will never succeed. Large Language Models will continue to improve, but despite this, they are not even *theoretically capable* of the forms of reasoning, knowledge management, and complex operations, which are required for serious real-world knowledge work (cf. Thórisson 2021; Thórisson, 2012).

Therefore, our response to the foundational paper of Conversational AI, "[Attention is All You Need](#)" is no, in fact, **Cognition is all you need.**

To reach more advanced levels of AI – for example, AI capable of meeting the demands of professional knowledge workers and knowledge organizations - we must innovate beyond the limits of the Conversational AI framework, and the Cognitive Layer is the best way forward for that agenda.

For experts in AI, venture capital, and technology trends, the coming shift to Cognitive AI signals a critical phase transition. For one thing, it means that investment into LLMs or similar-level alternatives, is likely to yield short term impressive gains, but diminishing long-term rewards, while the greatest potential future reward will come from investment into innovations and applications at the Cognitive Layer.

In other words, it would be wiser to invest in the Cognitive Layer instead of the Conversational layer, at this point in the innovation curve of both approaches. This requires a reassessment of current investment strategies in AI technologies, and a reevaluation of the potential applications and implications of AI across sectors.

*The next wave of AI is Cognitive AI.*

In this paper we will delve deeply into the arguments that prove this point, as well as their implications. Our arguments indicate that LLMs are a necessary but *insufficient* ingredient for complex knowledge work, while in contrast, CognitiveAI is *both necessary and sufficient*. The transition to Cognitive AI is inevitable and has already started.

# Related Research

We begin by exploring the limits of Large-Language Models, and the corresponding paradigm of Conversational AI, for meeting the needs of mainstream adopter knowledge workers.

Conversational AI is a necessary ingredient for applying AI to knowledge work, but it is not sufficient for the full set of needs that knowledge workers have. While LLMs may improve certain aspects of knowledge work productivity – such as speed of work – they do not necessarily improve the quality of knowledge work.

The underlying reason for this lies in Conversational AI's lack of actual cognitive processing, which limits the quality of insights it can deliver. We will explore cognitive processing in more detail in later sections of this paper, but first we examine evidence that indicates the insufficiency of LLMs for knowledge work.

Large language models have been widely celebrated for their remarkable performance across various natural language tasks, demonstrating the ability to achieve human-level performance on a wide spectrum of tasks (Moiseev et al., 2022). These models have been shown to encode substantial amounts of world and commonsense knowledge in their parameters, sparking significant interest in methods for extracting this knowledge (Haviv et al., 2021).

However, evidence suggests that large language models (LLMs) may enhance productivity but not necessarily improve the quality of work for professionals. While Devlin et al. (2019) demonstrated that scaling to extreme model sizes leads to significant improvements on small-scale tasks, indicating potential productivity gains (Devlin et al., 2019), in contrast Conneau et al. (2020) have highlighted that pre-training on Wikipedia, a relatively limited scale data set, may not sufficiently address the quality aspect, especially for lower resource languages (Conneau et al., 2020). This indicates that while LLMs may enhance productivity, the quality of work, particularly in diverse linguistic contexts, may not be significantly improved, unless models are extremely large.

Large Language Models have demonstrated exceptional performance in various natural language processing tasks and have shown the ability to solve reasoning problems (Ishay et al., 2023). However, LLMs face limitations in logical reasoning, which restrict their applicability in critical domains such as law (Nguyen, 2023). Existing literature exposes several challenges that LLMs face, including their lack of multi-step reasoning capabilities (Tongshuang et al., 2021), limitations in answering neurophysiology questions, and performing complex reasoning tasks (Shojaee-Mend, 2023). LLMs also lack transparency and explainability, making it challenging to obtain a complete picture of the knowledge reflected in a model or the reasoning used to produce its output (Liao et al., 2023). Moreover, the prospect of auditing LLMs is limited, and there are challenges in auditing LLMs at all (Mökander et al., 2023; Thórisson, 2021).

Beyond the limitations that stem from model size, and reasoning limits, recent research has also highlighted the limitations of large language models in capturing and utilizing knowledge effectively for serious knowledge work. For instance, it has been observed that large pretrained language models only learn attested physical knowledge, indicating a limitation in their ability to capture and utilize diverse forms of knowledge (Porada et al., 2019). Furthermore, while these models have shown impressive few-shot results on a wide range of tasks, they struggle with compositional generalization to novel examples, which is a crucial capability for serious knowledge work (Yang et al., 2022).

Moreover, the insufficiency of large language models for serious knowledge work is further underscored by their limited ability to reason and generate natural language proofs, as they struggle with reasoning in natural language and compositional generalization to novel examples (Yang et al., 2022).

Additionally, the challenge of adapting large parametric language models to evolving world knowledge without expensive model re-training further highlights their limitations in serious knowledge work (Pan et al., 2022). Furthermore, the fact that these models are trained on plain texts without introducing knowledge such as linguistic and world knowledge also points to their insufficiency for serious knowledge work (Sun, 2021).

While large language models have demonstrated impressive performance across various natural language tasks and have been shown to encode substantial amounts of world and commonsense knowledge, their limitations in capturing diverse forms of knowledge, reasoning, and adapting to evolving world knowledge underscore their insufficiency for serious knowledge work.

A study published in September 2023 by Harvard Business School showed that on average Conversational AI improved the work-quality of lower performers and sped up work in general, leading to better results approximately 40% of the time. However, to achieve mainstream adoption, it is necessary to innovate on the work-quality dimension.

A seminal study, "Navigating the Jagged Technological Frontier: Field Experimental Evidence of the Effects of AI on Knowledge Worker Productivity and Quality," (Dell'Acqua et al, 2023) conducted by Harvard University and BCG researchers examined the nuanced impact of AI on workforce productivity and accuracy, revealing a complex landscape where AI's benefits are accompanied by notable pitfalls.

While AI significantly boosted efficiency, enabling consultants to work faster, it also increased the likelihood of errors in tasks beyond AI's enhancement scope by 19 percentage points. In this experiment, BCG employees completed a consulting task with help from an LLM-powered chatbot. The bottom-half of subjects, in terms of skills, benefited the most, showing a 43% improvement in performance, compared to the top half whose performance increased by 17%. This finding underscores the necessity of preparing the workforce for the "jagged technological frontier" of AI—areas where AI excels versus where its application may lead to suboptimal outcomes.

The research suggests that while AI can dramatically improve operational speed and facilitate multitasking, it falls short in handling complex issues that demand human empathy and nuanced understanding. This dichotomy emphasizes the importance of strategic AI integration, where technology complements rather than supplants human capabilities. For organizational leaders, the study advocates for a balanced approach to AI integration, focusing on continuous learning and adaptation to AI advancements. It calls for a collaborative effort to harness AI's potential while mitigating its limitations, ensuring that AI and human collaboration synergize to propel innovation and success, avoiding the metaphorical "coffee-flavored jellybeans" scenario of unexpected and undesirable outcomes.

Similarly, other studies (Noy, S. et al 2023) found that people complete simulated information work tasks much faster and with a higher quality of output when using generative AI-based tools, however for some tasks, increased speed can come with moderately lower correctness (Spathariotiet al., 2023).

Another study, Microsoft's "AI and Productivity Report," cites multiple studies using Microsoft 365 Copilot observing information worker tasks for which LLMs are most likely to provide significant value (Cambon et al., 2023), in which most subjects agreed that Copilot helped them complete tasks faster, and the majority said it would help them get to a good first draft faster. However, several studies found no statistically significant or meaningful effect on work quality, despite subjects self-reporting the perception of improved quality.

In a study of a staggered rollout of a generative AI-based conversational assistant, Brynjolfsson et al. (2023) found that the tool helped novice and low-skilled workers the most. They found suggestive evidence that the AI helped disseminate tacit knowledge that experienced and high-skilled workers already had. In a lab experiment, participants who scored poorly on their first writing task improved more when given access to ChatGPT than those with high scores on the initial task (Noy & Zhang 2023). Peng et al. (2023) also found suggestive evidence that GitHub Copilot was more helpful to developers with less experience. Recent work by Haslberger et al. (2023) highlights further complexities and nuances in these trends.

In another relevant study by Choi et al., 2023, researchers conducted the first randomized controlled trial to examine the impact of AI, specifically GPT-4, on human legal analysis. Law students were assigned to complete legal tasks with or without GPT-4 assistance, with their performance speed and quality blind-graded. This study revealed that GPT-4 marginally improved the quality of legal analysis, notably among the least skilled participants, while significantly enhancing task completion speed for all. Participants reported greater satisfaction when using AI and identified tasks where GPT-4 was most beneficial. These findings suggest AI's potential to boost productivity, satisfaction, and even promote equality within the legal profession.

From the above cited research we conclude that while Conversational AI and Large Language Models (LLMs) offer substantial benefits in terms of speed and efficiency in knowledge work, their contribution to enhancing the quality of knowledge work remains questionable, due in part to limitations in their reasoning, logical analysis, and adaptation to evolving knowledge landscapes.

The integration of Cognitive AI into the knowledge workforce offers a path forward, where AI can not only improve knowledge work productivity, but also knowledge work quality. This represents a pivotal shift towards leveraging AI's strengths while effectively addressing the shortcomings of LLMs. In addition, by fostering a symbiotic relationship between human intelligence and AI's computational power, Cognitive AI can unlock new frontiers of collaborative innovation in knowledge work.

In the sections below we will conduct technical and theoretical comparison of Conversational AI versus Cognitive AI, for the purpose of knowledge work. We will show that LLMs are neither practically or theoretically capable of meeting the needs of knowledge work. While they may contribute to knowledge work by simulating aspects of these cognitive processes, these simulations have inherent limitations that cannot be overcome. The solution we propose is Cognitive AI, which is a new evolution of AI that performs higher-level cognitive processing, by harnessing the benefits of LLMs without being limited by their weaknesses.

# Defining Conversational AI

*Conversational AI* is a form of artificial intelligence based on conversations between agents. Here agents can be software agents or human agents. To be more precise, Conversational AI is a form of AI based on conversations which include at least two agents, where one is a software agent.

In Conversational AI agents communicate through streams of tokens, using Large Language Models (LLMs) to mediate their interactions. LLMs use underlying probabilistic models to generate token strings in response to token strings.

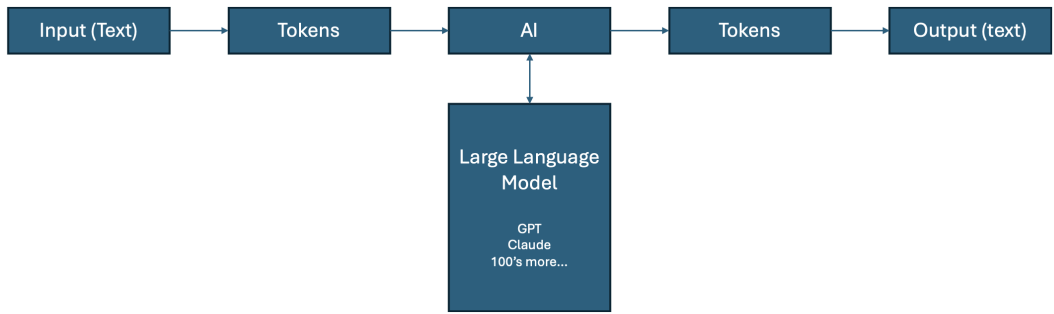

**Figure 1. Token Streams**

The fundamental units of Conversational AI are conversations or *chats*, which are streams of messages between agents.

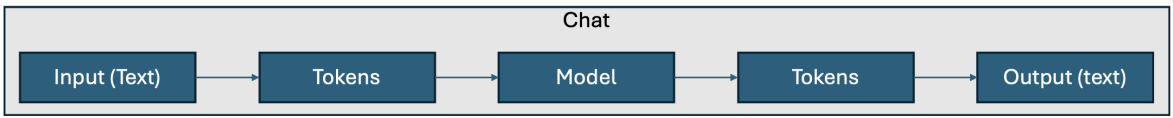

**Figure 2. Chats**

The user-facing manifestation of Conversational AI is manifest as a "*chatbot*," which is an application that executes a simple linguistic circuit between a software agent playing the role of the "bot" and a human user (or another software agent) that communicates with it.

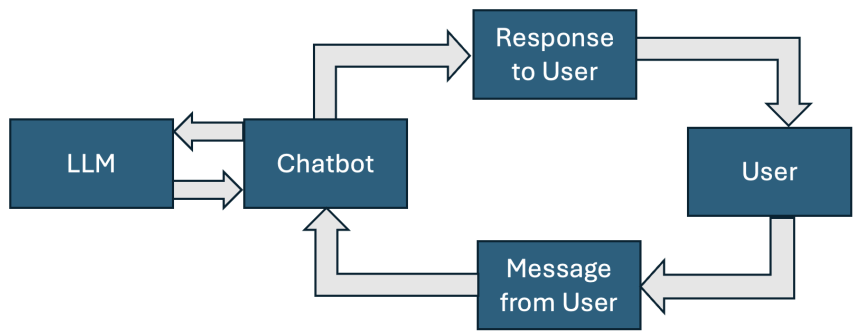

**Figure 3. Chatbots.**

By adding additional components to these circuits, they can make use of external data in the form of vector embeddings, and queries against vector databases, to augment the training of the underlying model at runtime. This makes these circuits able to incorporate new information that is not in the original training of the underlying LLMs.

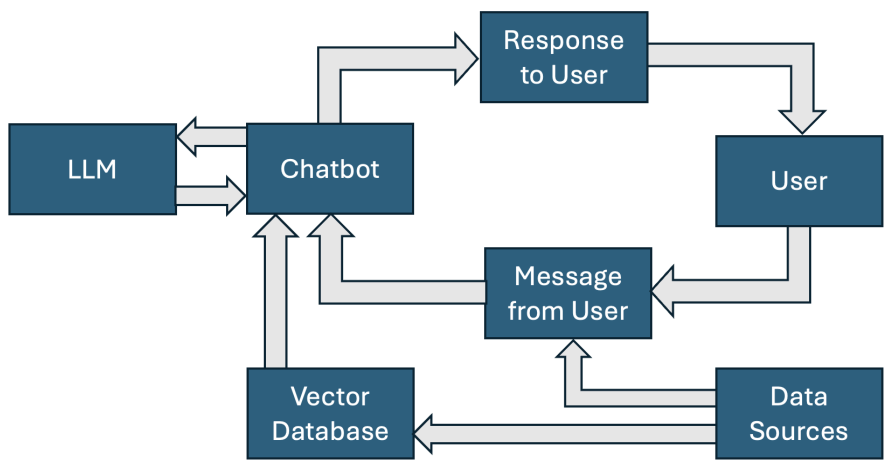

**Figure 4. Retrieval Augmented Generation (RAG).**

It is also possible to create multi-agent systems in which LLM-powered agents can engage in dialogs with each other (and optionally also with humans).

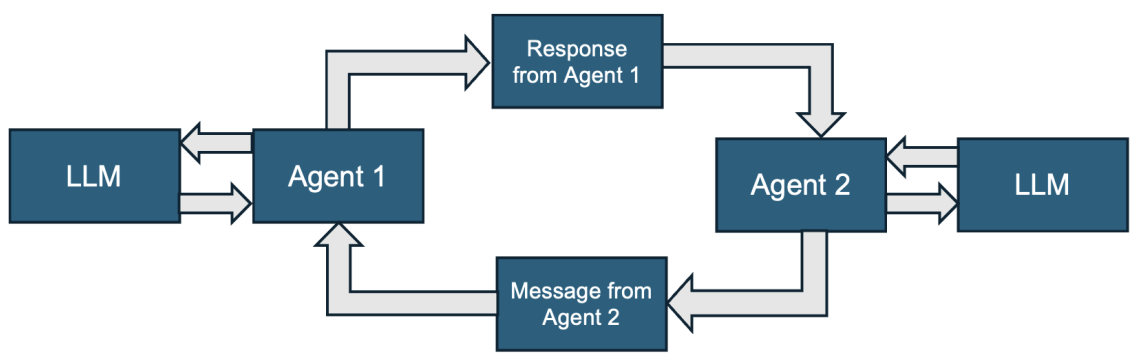

**Figure 5. Multi-agent dialogs.**

The process of Conversational AI, using state-of-the-art foundation models such as OpenAI's GPT 4 plus vector embeddings, enables a surprisingly powerful level of interactive artificial intelligence which is capable of answering questions and generating useful content about an infinite range of topics and data.

However, while Conversational AI achieves virtually unlimited breadth on simple tasks, the depth of its intelligence on more complex tasks is limited.

Conversational AI is a form of *"first-order intelligence"* that generates responses using LLMs, *without understanding, reasoning or reflecting on anything* (Thórisson et al., 2016)

Using *"prompts,"* and *"prompt-engineering"* methodologies it is possible to use language to guide the behavior of LLMs, in order to cause them to generate more specific outputs for various kinds of inputs. Prompts, like all messages between agents and the LLM, are saved to a "chat transcript" for an interaction session.

The *"chat transcript"* is a history of messages between agents, along with any added external information.

Chatbots operate with finite history. The *"context"* of a chat is defined as the set tokens that an LLM is given as input in order to generate a completion as output. Context cannot be longer than the maximum number of tokens an LLM can accept in a single input.
In a chatbot application, the interaction between agents and the LLM proceeds in a series of interleaved messages that constitute a "*dialog*." Messages, and dialogs composed of them, can be any length under whatever token length constraints are in effect.

The "token window" is the maximum number of tokens that can be provided as context to an LLM. In a dialog that produces a stream of tokens that exceeds the maximum number of tokens that the LLM can read in a single input, the token window is a moving window in the transcript, and is provided to the LLM as context for each input.

Using these basic constraints, Chatbots can generate dialogs that appear to be the products of intelligence and reasoning. However, in such dialogs only the subset of messages by human agents (such as a human user) involve any reasoning. Messages produced by the chatbots, which are generated by the LLM model, are in fact purely probabilistic streams of guesses which do not involve any understanding or reasoning.

It is a common misconception that chatbots understand what they say, or what users say, or what dialogs are about. In fact, for any given input such as a message from a human user, the chatbot simply uses the probabilistic weights in the underlying LLM to generate and return a stream of tokens that are correlated with the input above a certain probability threshold.

Instead of reasoning, Conversational AI applications generate statistical responses that seem to be the products of cognition, but are in fact only the products of *non-cognitive intelligence* that emerges from probabilities based on the numeric weights of the underlying models, which in turn are a consequence of their training and the data they were trained on.

Conversational AI is essentially a form of advanced mimicry of the cognitive processes, based on probabilistic models. Inherent in this fact are many inescapable built-in limitations which we will explore later in this paper.

# Intelligence Versus Cognition

*"Intelligence"* can be defined as the set of all systems that generate non-random output information in response to non-random input information. This is quite a broad definition, in which even physical processes and mathematical functions and formal systems can be considered to be forms of intelligence.

Within intelligence, the class of systems that are equivalent to Turing Machines conduct *computations*. Within the set of computations, *machine learning* systems exhibit the ability to make predictions based on learning. Likewise, computations that perform *artificial intelligence* generate outputs that more closely resemble those that humans can generate.

*"Cognition"* is a specific subset of intelligence, where the processing that systems do to transform inputs to outputs closely mirrors human cognitive processing. Within the scope of cognition, there are a number of critical cognitive processes that can take place, including learning and self-improvement, sensing, self-reflection and introspection, language understanding and processing, memory and context management, knowledge representation, knowledge management, knowledge processing, research and exploration, reasoning, planning, decision making, project management, and task execution.

LLMs produce outputs from inputs that seem to be the products of cognition. The linguistic (or visual, auditory, data) structures they generate are highly contextually relevant and appropriate responses to the meanings of the inputs they receive.

From a "black box" perspective - without knowing how LLMs work - one might assume they understand, reason, and even can be creative. However no cognitive activity is actually taking place within LLMs. They have no understanding of what is being said and they do not think, they merely process probabilities. However, despite this, LLMs produce surprisingly good responses that appear to be the products of cognitive processing, in other words they do a good job of mimicking cognition.

LLMs, and all Conversational AI systems, are classified as *intelligent, but not cognitive*, because they perform probabilistic natural language processing and response generation, but they do not actually perform higher level cognition.

# Instincts Versus Abstract Reasoning

The distinction between the operational mechanics of Large Language Models (LLMs) and the advanced functionalities of Cognitive AI highlights a fundamental divide between different forms of artificial intelligence: instinctual intelligence versus abstract cognitive reasoning. This divide not only characterizes the limitations and capabilities of these AI systems but also underscores the evolutionary trajectory from simple pattern recognition to complex cognitive processing.

LLMs operate on what can be described as "instinctual intelligence" in which responses are instinctual, meaning that they are provided automatically without any intermediate thinking or reasoning. Like instincts, which are innate, non-adaptive responses triggered by specific stimuli, LLMs respond to inputs based on patterns learned during their training phase. This process is inherently non-adaptive; LLMs cannot learn, reason, or change in real-time. Their responses, while sophisticated and often convincingly human-like, are limited by their training, lacking the capacity for live, on-the-fly learning or adaptation.

The interaction with an LLM, therefore, does not involve any genuine learning or memory integration. Responses generated during an LLM's operation are the result of processing input patterns against a static model, with no new information retained or integrated into the model's "knowledge" post-training. Even with the introduction of embeddings to augment LLM responses at runtime, the LLM processes these probabilistically, without engaging in actual learning or thought.

In stark contrast, Cognitive AI embodies the principles of abstract reasoning and second-order learning, engaging in a continuous loop of learning and adaptation even during runtime. Unlike the static, instinctual responses of LLMs, Cognitive AI's architecture allows for the accumulation of new knowledge, adjustment of strategies based on live feedback, and genuine reasoning about the content it processes. This dynamic capability enables Cognitive AI to not just simulate reasoning but to actually reason, learn from interactions, and evolve its understanding and responses over time.

Cognitive AI's approach to problem-solving and interaction is underpinned by structured knowledge and reasoning algorithms, facilitating a level of analysis, decision-making, and creativity far beyond the capabilities of LLMs. This not only allows for more accurate and contextually relevant responses but also supports the system's ability to engage in genuine abstract reasoning, drawing inferences, and generating hypotheses beyond the immediate input patterns.

It is expected that all of the major foundation models will continue to evolve and develop higher levels of world knowledge, comprehension, reasoning, user interaction, tool utilization, and self-improvement. However, these capabilities will still be mimicry as opposed to actual cognition.

While LLMs can offer powerful artificial intelligence capabilities through simulated reasoning, producing responses that are often surprisingly apt, the inherent limitations of this approach become apparent as the complexity of tasks increases. The inability to learn or adapt in real-time, coupled with a lack of genuine understanding or reasoning, places a ceiling on the intelligence that LLMs can achieve.

In contrast, Cognitive AI's capacity for abstract reasoning, continuous learning, and dynamic adaptation represents a significant leap towards overcoming these limitations, pointing the way towards more sophisticated, versatile, and genuinely intelligent AI systems.

The evolution from the instinctual intelligence of LLMs to the abstract cognitive reasoning capabilities of Cognitive AI marks a pivotal shift in artificial intelligence. By transcending the bounds of pattern-based responses and embracing the complexities of genuine learning and reasoning, Cognitive AI paves the way for AI systems that can engage more deeply with the world, solve more complex problems, and, ultimately, approach the elusive goal of Artificial General Intelligence. This shift from simulated reasoning to genuine cognitive processing defines the next frontier in AI, promising advancements that could redefine our understanding of what machines are capable of achieving.

# Defining Cognitive AI

*Cognitive AI* is a subset of artificial intelligence in which a cognitive layer executes *neuro-symbolic* cognitive processes that are modeled on individual and collective human cognition, by making use of a Cognitive Layer that uses a Conversational Layer.

The distinction between Conversational AI and Cognitive AI is precisely that Cognitive AI does not merely mimic cognition, rather it executes formal cognition outside of the underlying LLM models it uses. Therefore *Cognitive AI is classified as both intelligent and cognitive.*

By implementing the cognitive processes of the human mind, as well as collective intelligences of groups of humans, Cognitive AI is capable of self-directed thought and the orchestration of its cognitive processes, essentially enabling it to manage its knowledge work autonomously.

Cognitive AI transcends traditional AI's focus on pattern recognition and probabilistic predictions by incorporating a second layer of intelligence: meta-cognition. This advanced cognitive layer enables the AI to engage in genuine reasoning and learning from experiences, allowing for strategic adaptations in real-time. Such capabilities enable Cognitive AI to tackle complex, dynamically changing problems far beyond the reach of current LLMs.

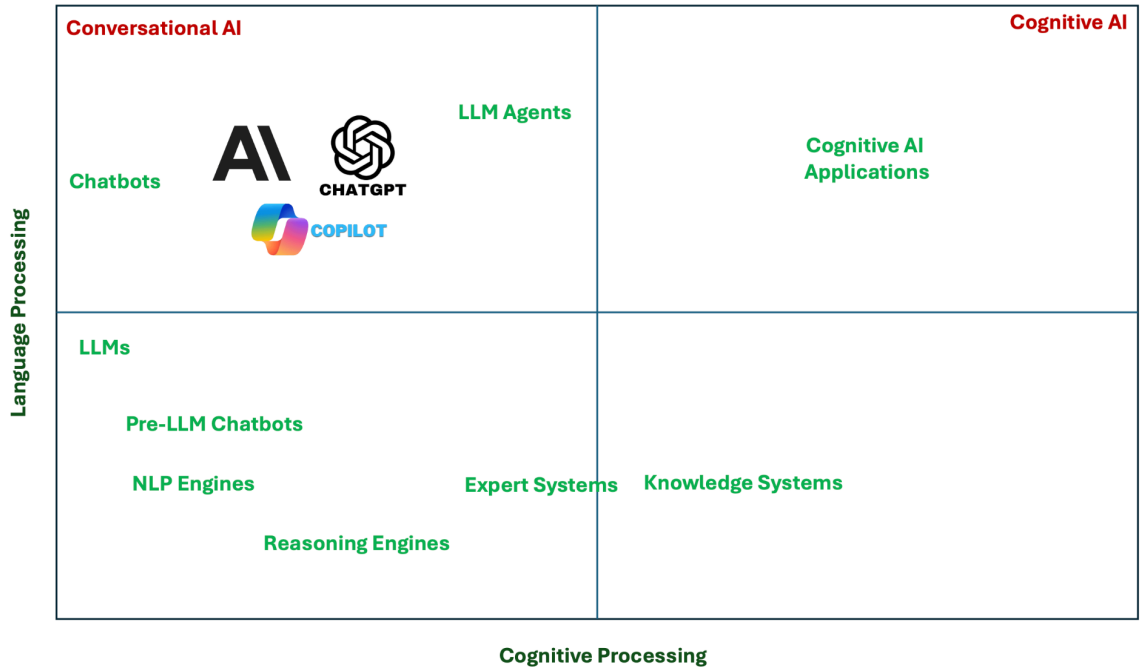

**Figure 6. Conversational Versus Cognitive AI Quadrants.**

# Cognitive AI Functional Architecture

Cognitive AI represents a paradigm shift, moving beyond the confines of Conversational AI's reliance on probabilistic reasoning simulations to actual programmatic reasoning. This shift is embodied in a dual-layer architecture that elevates reasoning, self-improvement, and adaptability to second-order intelligence, fundamentally distinguishing Cognitive AI from its predecessors. Below we will discuss the functional architecture and formal requirements for Cognitive AI systems.

## Functional Requirements for Cognitive AI

To qualify as Cognitive AI, a system must be architected to meet the following functional criteria:

1. **Dual-Layer Cognitive Architecture.** The system is organized into at least a dual-layer architecture, in which a Cognitive Layer that supports higher-level cognitive functions sits above a Conversational Layer that provides services equivalent to a general-purpose large language model.

2. **Large Language Models.** The system must provide and utilize one or more large language models (LLMS), or other similarly powerful and general alternative probabilistic models, in the Conversational layer, where at least one model is closely comparable to a large general purpose foundation model (such as GPT 4).
3. **Cognitive Agents.** The system must be architected with agentic design patterns and principles as an agentic application that provides intelligent cognitive agents which can operate semi-autonomously or fully autonomously, and where such agents are controlled and executed from outside of LLM transcripts, by an agent management function implemented as executable software.
4. **Relationship Management.** The system must enable the formation, management and use of social relationships to connect agents (including humans and software agents) on a one-to-one and one-to-many basis.
5. **Inter-Agent Messaging**. The system must enable natural language interactive messaging communication and the sharing of system objects (documents, knowledge, tools, data, agents, projects, plans, etc.) between agents that are directly or indirectly connected by mutual relationships.
6. **Dialogs**. The system must utilize interactive internal and external dialogs between two or more agents, where agents can be software-based or humans, and where in any dialog there is at least one software agent, and where dialog formats and execution can be structured and controlled with conditional logic rules..
7. **Planning**. The system must be able to generate, understand and respond to complex conditional workflows as plans in natural language.
8. **Project Management**. The system must provide a project management function to orchestrate and manage execution of plans by one or more agents.
9. **Neuro-Symbolic Reasoning**. The system must support generation of explicitly defined workflows for controlling both informal natural language reasoning and formal logical reasoning, where such workflows are executed and controlled from within the Cognitive Layer instead of from within the Conversational Layer.
10. **Memory Retrieval.** The system must be able to utilize its own planning and reasoning mechanisms to intelligently guide strategies for locating and retrieving relevant information and knowledge for a given context, across internal knowledge bases, long-term memory, and external data sources including the Internet.
11. **Context Management.** The system must manage context for agents and cognitive processes using a working memory to cache and swap relevant contextual information from long-term memory, in order to optimize relevancy of information in context against finite token windows of LLMs.
12. **Knowledge Discovery.** The system must conduct intelligently guided natural language and Boolean search, as well as deeper research strategies guided by agents (such as intelligently guided spidering for relevant information) to locate relevant information and knowledge across heterogeneous data sources (internal knowledge bases, long-term memory stores, and external resources including the Internet and third-party APIs).

13. **Knowledge Management.** The system must explicitly generate, learn, represent, store, retrieve and maintain formal data structures for representing knowledge that exist outside of LLMs.
14. **Tool-Utilization.** The system must have the ability to design and use tools in the form of software applications, APIs, and internal and external data sources. Tool-utilization also applies to a system being able to self-referentially utilize its own functional components as tools, to design and implement new tools, and to improve tools.
15. **Mathematics and Computation.** The system must have the ability to do mathematical and computational operations outside of LLMs, using software, data sets, and computing hardware and infrastructure. This also provides the system with advanced formal logical processing, scientific and financial calculation abilities, as well as data science and analytics and machine learning capabilities.
16. **Multi-Agent Collaboration.** The system must have the ability to orchestrate collaborative processes between human agents and software agents. This includes one-to-one, one-to-many, many-to-one, and many-to-many collaborative processes.
17. **Meta-Cognition.** The system must provide a meta-cognition function that can be utilized across all major cognitive functions of the system, and is capable of knowledge processing, introspection, meta-reasoning, reflection, learning, and self-optimization.
18. **Self-improvement.** The system must be able to engage in recursive goal-directed self-improvement, if and when needed, across all major cognitive processes, to iteratively optimize reasoning, knowledge, projects, plan, dialogs, agents, documents and code, both asynchronously and during runtime execution.

## Dual-Layer Architecture

At the core of Cognitive AI's functional architecture is an intelligence stack comprising two critical layers: a Cognitive Layer and a Conversational Layer.

The Conversational Layer operates on the principles familiar to LLMs (Large Language Models), processing and responding to linguistic inputs. Positioned above this, the Cognitive Layer introduces meta-cognition capabilities, extending the system's functionalities beyond mere linguistic processing to encompass higher-order cognitive processes.

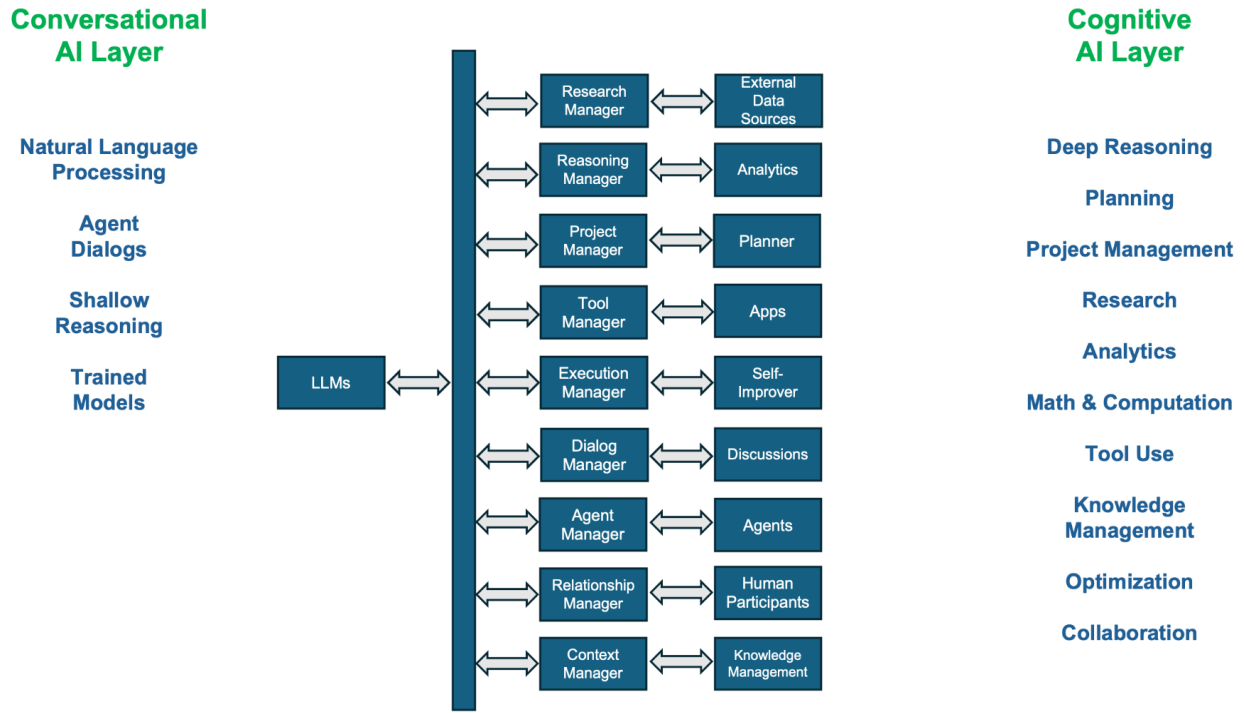

**Figure 7. Cognitive AI Functional Architecture**

The above diagram illustrates the functional architecture of Mindcorp's Cognition platform for Cognitive AI and can serve as a general model for how Cognitive AI architectures are designed.

The Cognitive Layer is where Cognitive AI truly differentiates itself. It provides the system with the ability to engage in meta-cognition (also called meta-cognition), in which it can engage in introspection, enabling a deeper level of understanding and optimization of its own processes. This functional area allows Cognitive AI to critically assess its methodologies for learning, reasoning, planning, and decision-making, mirroring the cognitive functions of the human mind engaged in complex knowledge work.

Through meta-cognition, Cognitive AI can refine and adjust its strategies at runtime, responding dynamically to new information and challenges. This adaptability is crucial for applications requiring not just an understanding of data but also the capacity to apply strategic thinking and creativity to solve problems.

The integration of meta-cognition equips Cognitive AI systems with the unique ability to self-assess their thought processes and learn from their interactions. This self-assessment capability ensures that Cognitive AI can continually refine its operational strategies, enhancing its efficiency and effectiveness over time. By continuously learning from its actions and the outcomes of its decisions, Cognitive AI can evolve its approach to problem-solving, ensuring that it remains effective in the face of changing conditions and requirements.

The architectural distinction of Cognitive AI, characterized by its dual-layer approach in which meta-cognition plays an important role, marks a significant advancement in the field of artificial intelligence. This cognitive structure not only enables Cognitive AI to process information linguistically but also empowers it with the ability to reason, plan, and improve itself autonomously.

By mirroring the cognitive processes of the human mind and incorporating the capacity for self-reflection and adaptation, Cognitive AI opens new avenues for solving complex problems, making it a powerful tool for real-world knowledge work and beyond. This architectural innovation lays the foundation for a new generation of AI systems capable of more sophisticated, adaptable, and effective problem-solving strategies, setting Cognitive AI apart from traditional Conversational AI technologies.

## Large Language Models

Large Language Models (LLM's) are a class of probabilistic language models, generally based on an attention-based transformer algorithm for predicting next tokens from a stream of previous tokens. These models are trained to make predictions that correspond to the knowledge inherent in the training data sets and fine-tunings may also be added.

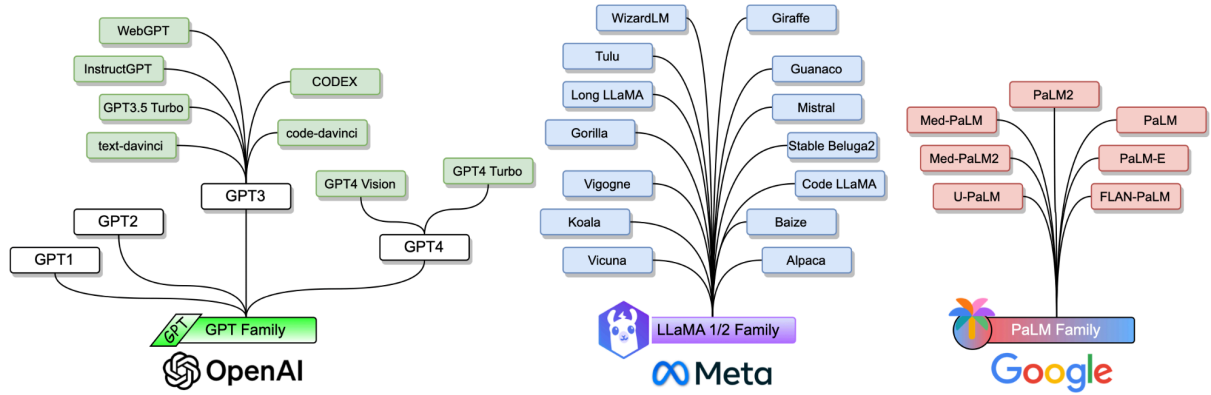

Figure 8. Leading LLM Models. Source: Minaee, S., et al., (2024).

The large foundation-level LLMs generate streams of tokens that contain sophisticated linguistic responses to streams of tokens. These responses are so similar to the kinds of intelligent responses that humans can generate, that the underlying LLM's are also said to be highly intelligent. However, as this paper will make exceedingly clear, there is a difference between intelligence and cognition. LLMs may be highly intelligent, but they are not cognitive at all, and this ultimately is their weakness.

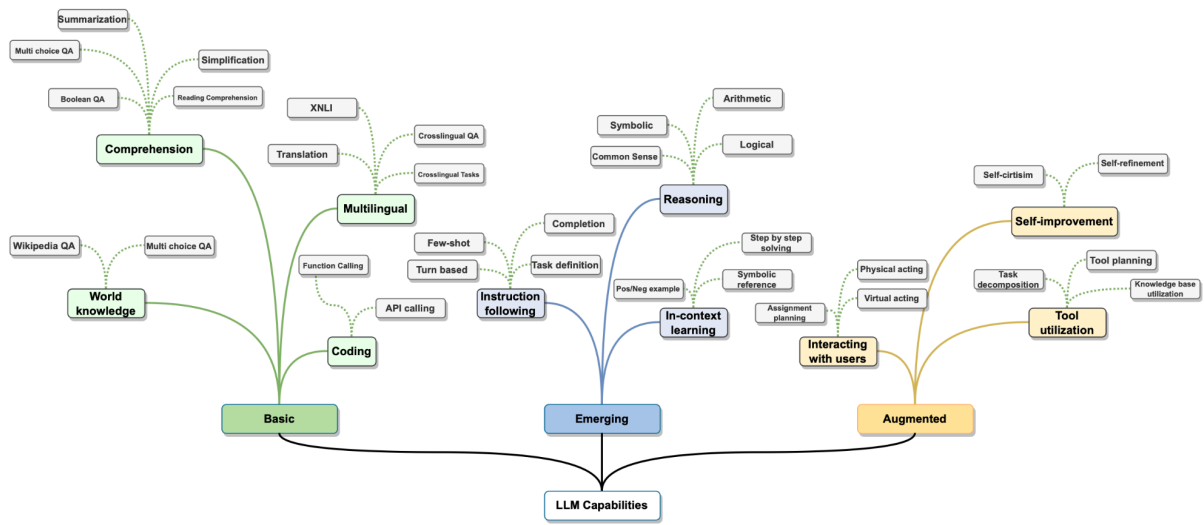

**Figure 9. LLM Capabilities. Source: Minaee, S., et al., (2024).**

The above diagram illustrates the current and emerging state of LLM capabilities. The diagram shows the capabilities of LLMs projected to be advancing into comprehension, reasoning, tool-utilization, social interactions, and self-improvement.

However it is important to note, and one of the main points of this paper, that within the context of LLMs, there is no actual comprehension, reasoning, tool-utilization, social interactions or self-improvement taking place. While the LLMs are very good at mimicking these processes to participate in simple conversations and generate basic documents, their ability to do so is shallow and falls apart quickly when problems get longer and/or multilayered and complex.

Despite the limitations of LLMs, they are the critical prerequisite for Cognitive AI. The language-level intelligence of LLMs is indispensable for Cognitive AI systems to function. However, in Cognitive AI systems the LLMs are not used to implement complex reasoning directly using language; instead complex reasoning and procedures are implemented on the Cognitive Layer, above the LLMs.

## Cognitive Agents

The genesis of Cognitive AI can be traced back to the burgeoning interest in agentic applications that operate above the LLM layer. Chatbots are the most widely-known example of the agent paradigm in the context of AI, but there are many other kinds of agents that can use LLMs without necessarily chatting or communicating with end-users.

These applications, characterized by their ability to complete tasks via one or more LLM-powered agents collaborating with a human and/or even with one another, mark the first steps towards transcending the limitations of LLM-driven models.

While initial forays into agentic AI have been focused on relatively simple tasks such as chatbots and various iterations of agents built on them, including agents that conduct online research, engage in social media, and complete simple online tasks, they lay the groundwork for more sophisticated, intelligent systems capable of complex decision-making and problem-solving.

As Cognitive AI begins to emerge, new agentic architectures and applications are forming above the LLM layer that do conduct rudimentary cognitive processing. In these platforms and applications, multi-agent systems provide agents that collaborate and/or compete to solve problems, using LLMs to think and converse within procedures that guide and channel this activity to perform forms of cognition.

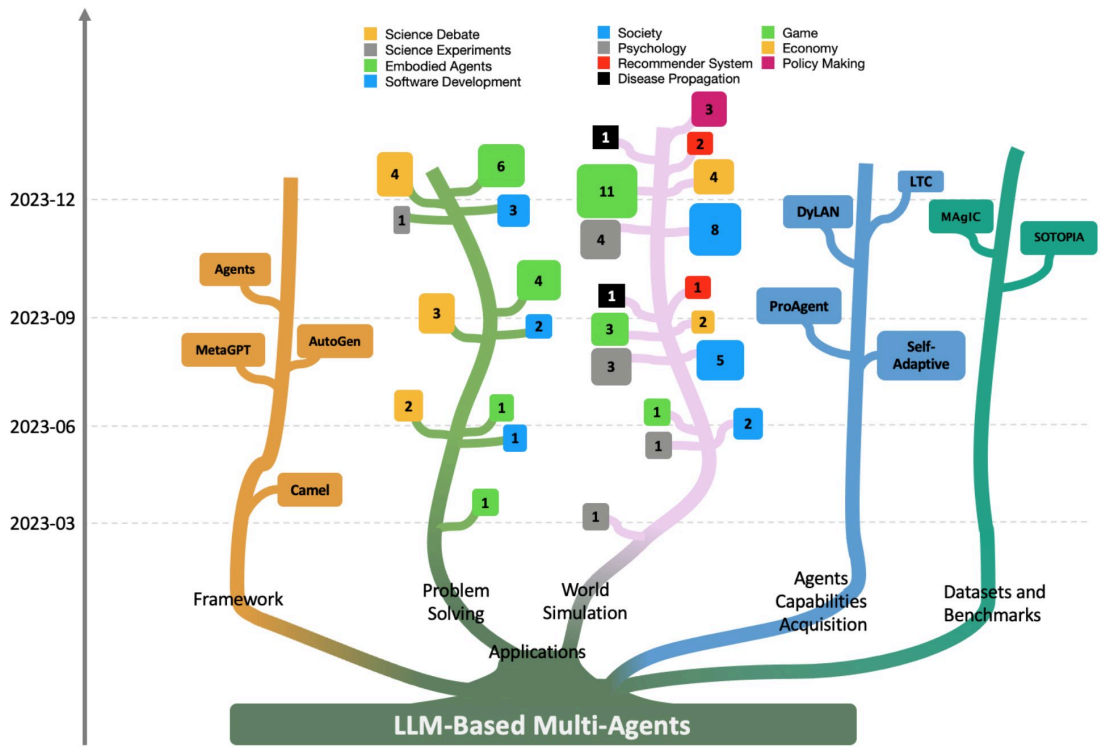

Figure 10. LLM-Based Multi-Agents. Source: Guo, Taicheng et al., 2024.

The diagram above illustrates the current state of the art in LLM-based multi-agent systems, where we observe a high degree of fragmentation across many competing approaches. This area of development is moving rapidly, but there is no common platform or any commonly accepted standards for agentic applications, inter-agent communication, or agents.

Below we illustrate what a complex multi-agent application might look like, at a high-level, for an example agentic "IP monetization" application:

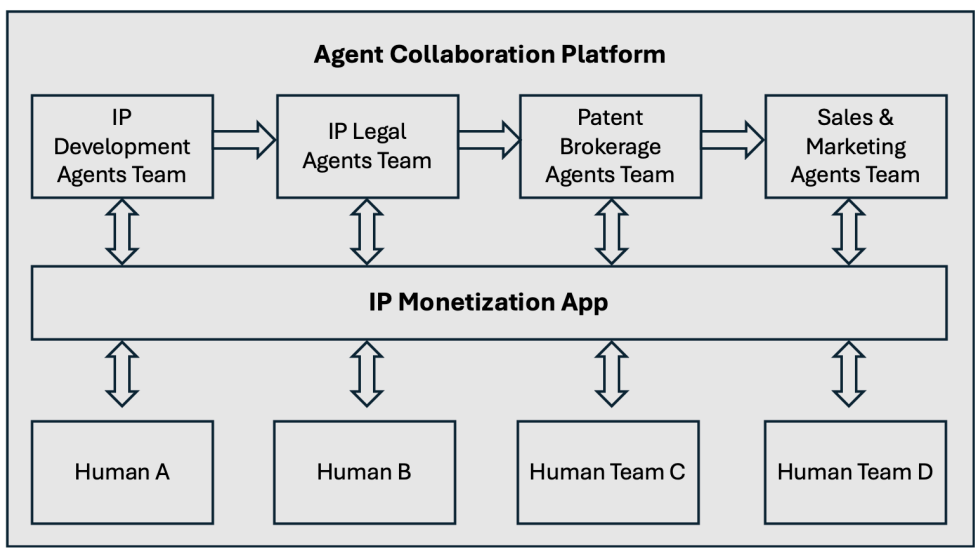

Figure 11. Agent Collaboration Platform

In the above example, several teams of agents collaborate with individual humans and teams of hums to conduct an IP development process. This is a highly advanced scenario. Most agentic LLM applications involve a single human delegating to multiple agents (single human, multiple agents: "SHMA"), for simple and relatively low-level task-automation scenarios like Web research. However in our own work (not yet released publicly, at time of this writing), we have implemented and tested a new agentic platform that is tailored for more advanced multi-human-multi-agent (multiple human, multiple agents: "MHMA") applications like this example.

Beneath this application are layers of modules, for example, the IP Development Agents team in the above diagram is a module that might function like the diagram below:

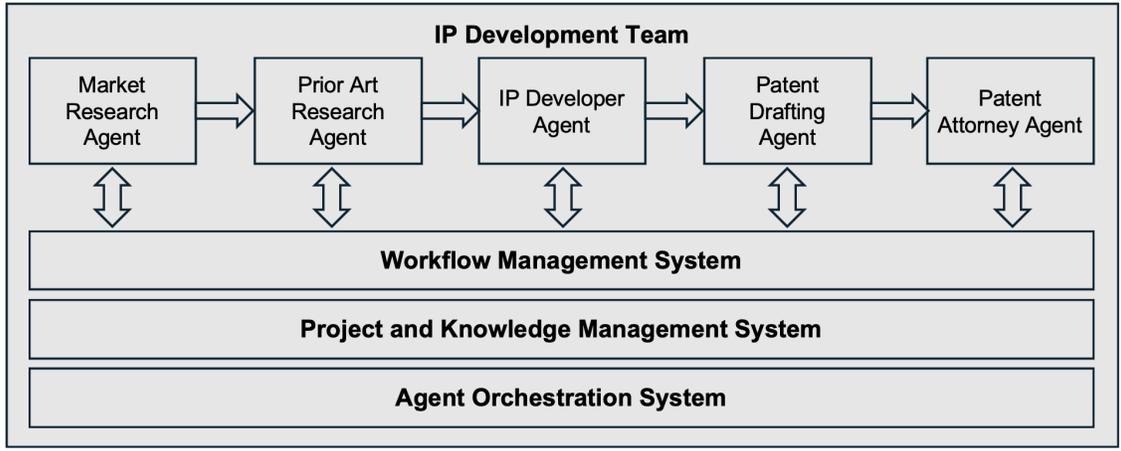

**Figure 12. IP Development Team Module.**

And below this level, each agent is a module - for example, the IP Developer Agent:

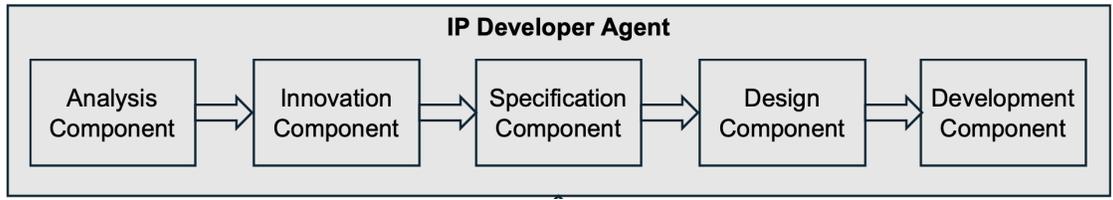

**Figure 13. IP Developer Agent**

And below this layer there are skills or sub-capabilities of each agent, for example:

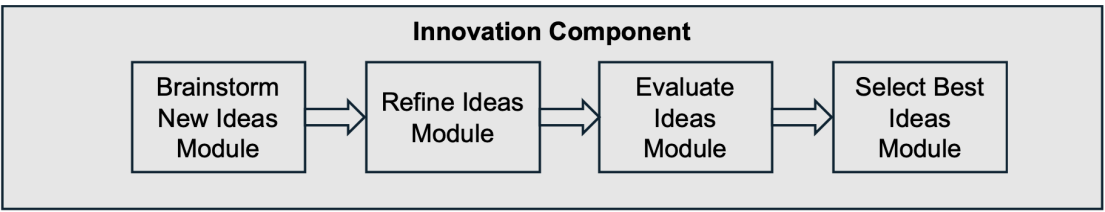

**Figure 14. Innovation Component**

But while agentic architectures are highly modular and well-suited to leveraging the capabilities of Large Language Models (LLMs), not all agentic applications rise to the level of full Cognitive AI.

*Only agentic applications, where agents are implemented outside of LLM transcripts, and where meta-cognition is utilized across all major cognitive functions, qualify as Cognitive AI.*

While LLMs themselves can mimic the behavior of individual agents and communities of agents, there is a difference between linguistically simulated agents that exist only as language within the transcript of an LLM chat, and programmatic intelligent agents that operate as executable code on a multi-agent platform above and outside of a chat transcript. The latter not only has more degrees of freedom in how they think and interact, but they can also make use of external code.

In the paradigm of Cognitive AI, agents exist outside of LLMs, but use LLMs for their linguistic and basic intelligence functions. In this way they utilize, but are not bound by, the limitations of LLMs. They are free, for example, to use multiple models, including other forms of machine learning and reasoning when needed.

The coming technical breakthroughs that enable Cognitive AI will have profound implications for the deployment of AI across various sectors. By enabling systems to understand and adapt their strategies, Cognitive AI will open new possibilities for innovation and problem-solving.

These adaptive capabilities ensure that Cognitive AI applications will continue to evolve and will remain relevant and effective in the face of changing data landscapes and problem sets, setting a new standard for what AI can achieve.

## Relationship Management, Inter-Agent Messaging and Dialogs

Cognitive AI thinks and reasons on at least two layers at once:
- The Conversational Level simulates cognition linguistically against a trained probabilistic model.
- The execution of explicit reasoning projects and plans by agents on the Cognitive Level constitutes programmable workflows that can model any informal or formal cognitive process.

By combining these two modalities a more powerful and flexible level of cognitive processing is possible than can be achieved by either on its own.

Where these two modalities intersect is in the process of dialogs. In Cognitive AI, dialogs are turn-based natural language inter-agent messaging conversations which take place between two or more agents, in which at least one participant is a software agent. A necessary precondition for inter-agent messaging and dialogs is a means for agents to form and manage inter-agent relationships.

There are two kinds of dialogs: external and internal dialogs. External dialogs take place between an agent and other agents that are external to its own private cognitive workspace. For example, a chat between a CTO Agent and a CFO Agent. Internal dialogs take place inside the scope of an agent's private cognitive workspace.

An internal dialog is a conversational process between two or more agents which takes place within a single agent's mindsteam. Within the cognitive workspace of an agent, sub-agents can be instantiated as needed, to represent facets of the subconscious processing of that agent. This effectively enables an agent to "talk to itself" and "cognitate internally" in order to self-reflect before generating a response or behavior. This form of internal discourse is critical for fostering deeper critical thinking, introspection, and analysis, within intelligent agents and across Cognitive AI architectures.

Through internal dialogs, Cognitive AI agents can meticulously develop, assess, and refine strategies and plans by applying dialectical processes between "subconscious agents" within its own virtual mind. This reflective and dialectical process enables a system to critique its own thinking, identify potential improvements, and iterate on its strategies before it puts them into practice as a response to some stimulus. Responses of this nature are far beyond the instinctive responses of LLMs. Internal dialogs, while not always required, can be used to ensure that responses are not only well-considered but also optimized for effectiveness and efficiency, embodying a level of strategic foresight comparable to human intelligence.

The process of internal dialogs for reasoning and adaptation in Cognitive AI is cyclical, constituting a continuous loop of self-improvement. This loop enables the system to evolve its problem-solving methodologies over time, ensuring that its approaches are not only effective but also increasingly sophisticated. By engaging in this ongoing process of introspection and self-modification, Cognitive AI systems can achieve a dynamic state of growth and learning, mirroring the evolutionary nature of human cognitive development.

The ability of Cognitive AI to evolve new and improved problem-solving strategies through introspective self-dialog and self-optimization, is critical for applications that require more than mere computational power. This capacity for sophisticated understanding and strategic planning, akin to human cognitive abilities, allows Cognitive AI to tackle complex tasks with a depth and efficiency that surpass the capabilities of Conversational AI.

Another key function of dialogs is group conversations. In Cognitive AI, group conversations are structured by plans that serve as their agendas, and they are facilitated by at least one software agent, for a group of two or more other agents. Cognitive AI agents are able to generate and leverage best-practices group processes for a variety of collective cognition tasks such as brainstorming, content development, research and analysis, strategic planning, design and development, innovation, feedback and reporting, and decision-making.

One of the more powerful applications of group conversations is the use of groups of agents with diverse specializations and skills to model collaborative multi-disciplinary teams and their collective cognition. In Cognitive AI, the practice of applying multi-disciplinary teams of agents is a routinely used mechanism during execution of projects and plans.

For example, during a particular step of a plan in a market research project, a team of agents can be assembled to discuss a market segment, where each agent brings unique knowledge, heuristics, and skills to the table. The team can then engage in a structured conversation, where each agent represents its unique perspective, to arrive at a richer understanding together.

## Planning and Project Management

LLMs have been shown to have limited planning capabilities and in recent benchmarks they still have much room for improvement. (Valmeekam, K. et al., 2023). LLMs also fail at over-the-horizon reasoning, where there are long complex chains of potential solutions, only some of which are optimal or even solutions at all.

Kambhampati, S. et al. have argued persuasively that "LLM's can't plan" because, for example, LLMs can neither guarantee the generation of correct plans, nor the verification of correct plans. Planning with LLMs is not equivalent to exhaustively searching for valid optimal paths in a solution space, but instead is more like generating plans by borrowing from previously seen plans – an approach which is not systematic.

The ability to strategize and plan thinking processes underpins a wide range of capabilities critical to Cognitive AI, including reasoning, research, analytics, decision-making, project management, and task orchestration. By embedding formal planning capabilities into the fabric of Cognitive AI's operations, these systems can tackle sophisticated challenges that require not only raw computational power but also nuanced, strategic thinking.

At the core of Cognitive AI's planning function is the ability to generate plans which are formal conditional workflows for agents to participate in. These workflows guide the collective cognition and behavior of intelligent agents, and optionally human collaborators as well, by channeling their interactions and reasoning through structured processes that guide them towards goals. This capability is essential for orchestrating the efforts of multiple entities, ensuring that each contributes effectively to the task at hand, based on their unique strengths and capabilities.

More specifically the plans generated by Cognitive AI may include formally specified plans, using a formal plan reasoning language such as PDDL. By integrating PDDL, or languages like it, into Cognitive AI systems, it becomes possible to conduct formal search, analysis, validation and optimization of plans, against formally specified problem domains, using first order predicate logic.

By combining this level of formal reasoning about plans with the informal language understanding and generation of the LLMs, a more sophisticated form of plan generation and refinement becomes possible, where the LLM generates potential plans with natural language, which are then transformed into formal logic, and which are next formally evaluated and improved, in order to yield better plans, with are finally translated back into natural language.

The plans developed by Cognitive AI systems are not rigid scripts but adaptive strategies that respond to changing conditions and new information. By programmatically channeling the collective thinking processes of teams of agents and humans, Cognitive AI can navigate complex problem spaces with agility and precision. This approach allows for the optimization of cognitive resources, ensuring that tasks are approached in the most efficient and effective manner possible.

Supplementing its planning capabilities, Cognitive AI incorporates full project management functionalities. Project management allows a system to not only devise and initiate plans, but also to monitor their progress, adjust execution at runtime, and manage resources effectively. Through comprehensive project management, Cognitive AI can engage in complex, multi-step, and long-term or ongoing knowledge work.

This integration of planning and project management enables Cognitive AI to orchestrate complex endeavors, from initial strategy formulation to the successful completion of objectives. It represents a holistic approach to tackling knowledge work, where the system's cognitive functions are leveraged to plan, manage, and execute projects with a level of sophistication and adaptability previously unattainable.

The planning and project management capabilities of Cognitive AI mark a significant advancement in artificial intelligence. By enabling dynamic, real-time strategizing, planning and execution, supplemented with comprehensive project management tools, Cognitive AI systems can effectively orchestrate complex collaborative knowledge work and knowledge-based business processes. This not only enhances the efficiency and effectiveness of cognitive work but also expands the possibilities for what AI can achieve, setting a new standard for intelligence in technology.

Cognitive AI's ability to integrate meta-cognition across planning and project management also enables a higher level of control and sophistication in reasoning and adaptation that is essential for complex problem-solving and knowledge work. Complex reasoning in Cognitive AI is the result of applying systems of agents to solve abstract problems, using projects and plans to do so. In other words, in Cognitive AI, complex reasoning is a cognitive process that uses projects and planning to control and manage multi-agent reasoning and behavior.

In Cognitive AI, planning and project management are central cognitive functions. These advanced AI capabilities are not only about creating, executing, and adapting strategies and plans in a static sense, but dynamically doing so in real time, as agents engage in cognitive processes. This dynamic planning capability is fundamental to the ability of agents to navigate and manage complex tasks, embodying a leap beyond traditional AI's capabilities.

## Neuro-Symbolic Reasoning

It has been argued above that LLM's are not capable of complex reasoning, however even naive logical reasoning within LLMs is prone to failures. For example, they are prone to the "reversal curse" (Berglund, L., et al., 2023) where if told that "A is B", they may fail to infer that "B is A", and in addition they often fail on even simple set operations, such as three set logical unions (Yang, J. et al., 2023).

For larger, more complex reasoning processes that are longer than the context window or token limit, LLMs cannot natively mimic complex reasoning because they cannot maintain context or state beyond the limits of their token limits. Furthermore, because of their proactive nature, LLMs cannot reliably implement complex reasoning in a deterministic manner; they are prone to hallucinations and unpredictable outputs and they may or may not always follow plans they are given.

Cognitive AI addresses these challenges by offering a neuro-symbolic solution that combines deterministic and programmatic control of reasoning and planning, which are executed outside the token window of an LLM, with the non-deterministic, probabilistic conversations that take place inside the token window.

Furthermore Cognitive AI can generate, validate, and optimize these external reasoning flows using formal symbolic processing and computation. In other words, Cognitive AI systems combine the "neuro" capabilities of LLMs with the "symbolic" capabilities of pre-LLM generations of AI, such as formal symbolic logic process, solvers, formal planners, formal reasoning engines and non-axiomatic reasoning methods (cf. Latapie et al., 2023).

This hybrid approach enables Cognitive AI systems to navigate complex reasoning tasks with greater precision and reliability. By integrating both deterministic and non-deterministic methodologies, Cognitive AI can leverage the strengths of each, resulting in a reasoning capability that is more powerful and versatile than either approach alone.

Unlike LLMs, which rely on language simulation to approximate thinking, Cognitive AI systems employ programmatically structured, organized, and managed thought processes. This programmatic control extends to the execution of projects and plans, allowing Cognitive AI to engage in complex autonomous reasoning. This structured approach to thinking and reasoning enables Cognitive AI systems to process and analyze information in a manner that aligns more closely with the requirements of sophisticated cognitive tasks.

Cognitive AI further enhances its reasoning capabilities by channeling conversational AI through agents that employ projects and plans to control their behaviors. This enables the collaborative and complex autonomous reasoning necessary for high-level knowledge work. By combining deterministic programmatic control with the flexibility and adaptability of non-deterministic conversational AI, Cognitive AI can tackle complex problems with both precision and creativity.

Another important aspect of reasoning in Cognitive AI is the capacity to construct and reason about formally defined systems of rules. These systems of logical rules can be processed with first-order predicate logic in symbolic processing modules, such as theorem provers, graph search algorithms, and reasoning engines.

Agents in Cognitive AI systems can execute, manage and improve, goal-directed projects and actions, under formal systems of rules. This enables such systems to intelligently, discover, reason about, and improve their own solution paths as they work, and adapt to change.

The reasoning capabilities of Cognitive AI represent a significant advancement over traditional conversational AI systems. Through the integration of reflection, planning, and programmatic control, Cognitive AI can navigate complex cognitive tasks with a level of sophistication and effectiveness unmatched by LLMs alone.

This approach to reasoning not only enhances the system's ability to perform complex problem-solving but also positions Cognitive AI as a critical tool for advancing knowledge work and other applications requiring nuanced, intelligent analysis and decision-making.

Cognitive AI's unique combination of deterministic and non-deterministic reasoning processes establishes a new benchmark for what artificial intelligence systems can achieve in terms of autonomous reasoning and cognitive collaboration.

## Memory Retrieval and Context Management

A defining feature of Cognitive AI, distinguishing it from Conversational AI, is its advanced capability for contextual understanding and memory, enriched by the integration of concepts akin to human working memory and long-term memory. This sophisticated memory system enables Cognitive AI to handle information dynamically and strategically, offering a substantial edge in complex problem-solving and nuanced decision-making.

Traditional Conversational AI processes each interaction in isolation, limiting its ability to recognize the continuity in ongoing dialogues or projects. Cognitive AI, however, boasts a contextual memory that spans interactions, acting as a dynamic repository of context, insights, and understanding. This system allows it to build upon previous conversations, adapting to context changes over time, and making more informed decisions and responses.

Central to how Cognitive AI handles memory is the process of context management, whereby it is able to provide relevant contextual information within finite token windows of LLMs. Context management is encapsulated in a working memory buffer that temporarily holds and manages information that is immediately relevant to the task at hand, akin to human working memory. This feature is crucial for maintaining the context of ongoing interactions, allowing for real-time analysis, reasoning, and the dynamic adjustment of strategies based on immediate feedback. Through working memory, Cognitive AI can juggle multiple pieces of information, drawing on recent data for predictions, problem-solving, or strategy adjustments.

Long-term memory in Cognitive AI mirrors the human capacity to store and retrieve information over extended periods. It enables the AI to accumulate knowledge and experiences, applying historical data to inform future decisions. This capability allows Cognitive AI to recognize patterns, learn from past interactions, and adapt its operational strategies, providing a depth of understanding and strategic foresight that traditional Conversational AI cannot achieve.

The dual-approach to memory—combining working memory and long-term memory—affords Cognitive AI a clear advantage over Conversational AI. While Conversational AI might access databases or follow scripted responses, it lacks the dynamic, context-aware integration of information that Cognitive AI offers. This system not only ensures more accurate, relevant responses in the moment but also enables deeper strategic thinking and learning over time.

Cognitive AI's memory system facilitates anticipation of needs, adaptation to contextual changes, and the delivery of solutions informed by both immediate and historical perspectives. This nuanced approach, mirroring human cognitive processes, allows Cognitive AI to navigate complex interactions and problem spaces with unprecedented depth of understanding and adaptability.

The integration of contextual understanding and memory, incorporating working memory and long-term memory, into Cognitive AI systems marks a significant evolution in artificial intelligence. This advancement allows Cognitive AI to operate with sophistication and adaptability, navigating complex challenges with strategic insight and efficiency. By leveraging both immediate and accumulated knowledge, Cognitive AI sets a new standard for AI-driven applications, offering sophisticated, adaptable, and effective solutions that far surpass the capabilities of traditional Conversational AI.

## Knowledge Discovery and Knowledge Management

A foundational element distinguishing Cognitive AI from conventional Conversational AI is its approach to knowledge representation and knowledge management (KM). This functional area transcends the capabilities of Large Language Models (LLMs) in handling knowledge, by explicitly creating, managing, and enhancing structured knowledge representations such as knowledge bases, knowledge objects, knowledge catalogs, knowledge graphs, taxonomies and ontologies. These knowledge representation components are pivotal for the sophisticated operation of Cognitive AI, enabling it to process, understand, and interact with information in a more nuanced and contextually relevant manner.

Unlike LLMs, which rely on pattern recognition and data-driven learning without an explicit structure for organizing knowledge, Cognitive AI systems are meticulously designed to construct and utilize knowledge representations, which are data structures not merely streams of natural language. The ability to process knowledge representations mirrors human cognitive processes, facilitating the AI's ability to draw connections, identify patterns, and efficiently access relevant knowledge.

Cognitive AI incorporates LLMs not as standalone entities but as integral tools that enrich the Knowledge Management process. LLMs are employed to understand, summarize, expand, and filter knowledge within Cognitive AI's knowledge bases. This symbiotic relationship between LLMs and Cognitive AI's structured Knowledge Management functions allows for the generation of rich metadata, the improvement of searchability, and the discovery and creation of new knowledge representations, which are processes that LLMs alone cannot replicate. By utilizing LLMs in this capacity, Cognitive AI systems can maintain a dynamic and ever-evolving knowledge base.

At the core of Cognitive AI's Knowledge Management is the function of knowledge discovery and improvement, a process that involves the active location, extraction and discovery of new insights based on accumulated information. As information (including raw data and information, and knowledge structures themselves) is added, additional knowledge can be extracted or inferred, including new metadata, new knowledge representations, and new relationships between existing representations. This function signifies Cognitive AI's ability to not only organize existing knowledge but also to continuously refine and expand its understanding through the integration of newly sourced data. The strategic use of LLMs in this context enhances Cognitive AI's capability for complex problem-solving and informed decision-making.

The capacity for research and information exploration marks another critical aspect of Knowledge Management in Cognitive AI. Beyond mere keyword matching, Cognitive AI systems deploy advanced algorithms to delve into queries against multimodal datasets, using their understanding of contexts, goals and intent, to retrieve contextually relevant information. This capability is crucial for tasks requiring domain-specific expertise or the navigation of vast data pools to uncover precise insights or solutions.

Knowledge Management within Cognitive AI architectures is a testament to the sophisticated nature of these systems, enabling them to effectively manage and leverage complex datasets and information structures far beyond the capabilities of LLMs. The integration of LLMs enhances Cognitive AI's ability to organize, access, and evolve its knowledge base, ensuring continuous improvement and relevance.

This comprehensive approach to Knowledge Management not only underscores Cognitive AI's transformative impact on artificial intelligence but also its indispensable role in facilitating complex decision-making, problem-solving, and innovation in various domains. Through its structured yet dynamic Knowledge Management system, Cognitive AI represents a significant leap forward in the pursuit of more intelligent, adaptable, and effective AI solutions.

## Tool-Utilization

Cognitive AI architectures are designed to support and integrate tool-utilization, which is the capacity for an intelligent system to use, and even create, tools . The cognitive approach to tool-utilization, also called tool-use, represents a significant point of differentiation between traditional conversational AI and the more advanced Cognitive AI systems. While conversational AI can simulate or script the control of tools, it lacks the inherent capacity to truly understand or directly interact with these tools. Cognitive AI, however, embodies tool-use as a core capability, reflecting a deeper layer of intelligence and functionality.

Conversational AI's approach to tool-use is essentially indirect. It can generate scripts or commands that control tools, such as computer programs, but any actual tool interaction must be executed externally. The AI itself remains disconnected from the tangible effects of these tools; it cannot perceive the tool or the results of its use directly. This limitation confines conversational AI to a role of an intermediary rather than an active and aware participant in tool-use.

In contrast, Cognitive AI is inherently designed for tool interaction. This design extends from the ability to reflexively use its own internal cognitive agents, processes, projects, and plans, treating them as tools within its operational framework. Furthermore, Cognitive AI's capacity for tool-use expands externally, enabling it to utilize a wide array of tools, from Large Language Models (LLMs) to various functions and APIs, effectively extending its capabilities and enhancing its operational efficiency.

It's particularly interesting to point out that Cognitive AI can intelligently use external data analytics and machine learning tools to analyze data, make predictions, and generate insights. This even extends to the ability for Cognitive AI to develop or fine-tune LLM models if necessary.

Cognitive AI can, if it needs software it cannot source from elsewhere, is capable of developing its own software tools. It can design, build and execute programs, create datasets, and even operate software-as-a-Service (SaaS) to meet the needs of its projects and tasks. AI code-generation and coding co-pilots already offer dramatic productivity gains for software engineers. However by using a cognitive layer above such tools, it becomes possible to go beyond them – Cognitive AI will be able to autonomously design, build, test, improve, operate and maintain SaaS applications.

The tool-use in Cognitive AI involves two critical aspects: the ability to interact with and control the tool and the ability to observe both the tool and the environment to make real-time adjustments. This dual capability allows Cognitive AI systems to engage in a form of meta-tool-use, where they not only utilize tools but also understand and optimize their use based on ongoing feedback and environmental conditions.

Unlike conversational AI, which operates on predefined instructions without the capacity to observe or adjust its actions, Cognitive AI systems possess the autonomy to manage tools actively. They can assess the effectiveness of a tool in real time, adapt their strategies to optimize its use, and even switch between tools as necessary to achieve their objectives. This level of operational adaptability and awareness enables Cognitive AI to perform complex tasks with a degree of precision and efficiency that conversational AI cannot match.

Cognitive AI's ability to utilize external tools, such as LLMs and APIs, signifies a substantial expansion of its operational domain. By integrating these external resources, Cognitive AI can enhance its problem-solving capabilities, access a broader range of information and functionalities, and execute tasks that are beyond the reach of its internal resources alone. This external tool-use not only amplifies the system's capabilities but also demonstrates the system's ability to operate within and contribute to a larger ecosystem of technologies and services.

Tool-use is not just an additional feature of Cognitive AI but a fundamental aspect of its design and functionality. It exemplifies the system's advanced level of intelligence, showcasing its ability to interact with, control, and adapt the use of both internal and external tools in the pursuit of its objectives.

Tool utilization distinguishes Cognitive AI from conversational AI, highlighting its potential to revolutionize how artificial intelligence systems engage with the world and accomplish tasks. Through integrated tool-use, Cognitive AI sets a new standard for autonomy, adaptability, and functionality in the AI domain, promising to redefine the boundaries of what artificial intelligence can achieve.

## Mathematics and Computation

A special area of tool utilization in Cognitive AI systems is the ability to use mathematical software and computing infrastructure as tools.

Cognitive AI systems must be capable of performing mathematical and computing operations, including writing software and then using that software to execute arbitrarily complex computations. They can also access and utilize any needed external IT infrastructure and data for which they have credentials. These capabilities do not come from the Conversational Layer, but instead are executed using dedicated code or applications for these purposes that either is built-into, or is called from, the Cognitive Layer.

By virtue of this, Cognitive AI systems are able to do arbitrary computations on an as-needed basis, and they can either use existing external tools and services for this, or they can develop software for tools they need.

Using mathematical and computational tools, Cognitive AI systems are capable of harnessing machine learning and data analytics, predictive analytics, financial analysis and modeling, formal reasoning, and mathematical and scientific computing.

## Multi-Agent Collaboration

Cognitive AI is inherently collaborative. The ability to interact, communicate and collaborate to conduct knowledge work with other individuals and groups is a key defining factor of higher-order intelligence on the evolutionary ladder.

Drawing inspiration from Marvin Minsky's "Society of Mind" approach, Cognitive AI's groundbreaking approach to artificial intelligence emphasizes the critical role of social collaboration in cognition, not just as a feature set but as a core mechanism driving many cognitive processes.

Cognitive AI systems enable collaboration between humans and software agents, as well as between software agents and other software agents, in natural language. This is a significant advance from previous pre-LLM agent paradigms where communication between agents was largely programmatic and incomprehensible to non-programmer humans.
In Cognitive AI, collaboration between humans and agents requires similar social infrastructure to collaboration between humans – there must be a way for them to message each other and communicate in dialogs, and there must be a way to organize their interactions into topics, projects and plans. The Project Manager function of a Cognitive AI architecture is the intersection where collaboration is organized, managed and executed.

Cognitive AI extends beyond human-AI interactions, encompassing the dynamics between agents, among individuals, and across groups that include both agents and people. By harnessing social interaction and collaboration, Cognitive AI taps into the power of collective intelligence, which is inherently social and collaborative at its core.

In Cognitive AI, both AI and human agents interact with one another to share knowledge, negotiate tasks, and collaboratively solve problems. These interactions are not merely transactional but can be built upon complex social and reporting mechanisms that allow agents to understand and predict each other's behaviors, seek appropriate approvals, work towards common goals, and optimize collective outcomes. This ability to engage in sophisticated social negotiations, reporting structures and collaborations enhances the agents' effectiveness and contributes to the development of a more nuanced and dynamic form of collective intelligence.

Cognitive AI also plays a pivotal role in facilitating and enriching human-to-human interaction. By offering insights derived from its processing capabilities and its understanding of social dynamics, Cognitive AI can facilitate group processes to help human teams communicate more effectively, brainstorm or innovate more effectively, overcome misunderstandings, and make decisions more efficiently. It acts as a mediator and enhancer of human collaboration, leveraging its understanding of group process and group dynamics to foster a more cohesive and productive working environment.

The interaction between Cognitive AI agents and humans is a hallmark of its social intelligence. Cognitive AI is designed to understand and adapt to human social cues, enabling it to participate in conversations and collaborative efforts as a proactive and responsive partner. This seamless integration of AI into human social contexts allows for a symbiotic relationship where both agents and humans learn from and complement each other's strengths, leading to enhanced creative problem-solving and innovation.

Perhaps most significantly, Cognitive AI facilitates and thrives in mixed groups comprising both AI agents and humans. In these settings, Cognitive AI leverages its social intelligence to navigate the complexities of group dynamics, ensuring that contributions from both AI agents and humans are valued and integrated. This collaborative environment maximizes the benefits of collective intelligence, where diverse perspectives and capabilities are brought together to tackle challenges more effectively than any individual or homogeneous group could.

At its core, Cognitive AI is built around the concept of collective intelligence, recognizing that the most profound insights and solutions often emerge from collaborative efforts. This collective intelligence is inherently social and collaborative, drawing on the diverse experiences and knowledge bases of its participants. By facilitating and actively participating in this collaborative process, Cognitive AI not only enhances its own cognitive capabilities but also contributes to the collective wisdom of the group.

The integration of social interaction into Cognitive AI represents a transformative advancement in AI technology. By embracing the complexities of social dynamics across various interaction modalities—agent-to-agent, human-to-human, agent-to-human, and in mixed groups—Cognitive AI elevates the potential for collective intelligence.

This approach acknowledges that the future of intelligence, both artificial and human, lies not just in individual brilliance but in our capacity to collaborate, share, and innovate together. Cognitive AI, with its emphasis on social interaction and collective intelligence, is leading the way toward a future where AI is an integral and enhancing component of the social fabric, propelling us towards more collaborative, intelligent, and creative outcomes.

# Meta-Cognition

Meta-cognition encompasses a particular set of advanced functions within an intelligent system, whereby the system is able to sense and react to itself through a process of self-reflective introspective reasoning.

In self-reflective reasoning, the thinking and behavior (including both the internal state and the output) of a meta-cognitive system is sensed by that system, and taken as input for further intelligent processing by that same system about itself. In other worlds, the system thinks about itself and what it is doing in order to improve its composition or behavior.

Implementing meta-cognition across agentic applications is technically difficult and is not a fully-implemented construct in existing agent development frameworks. While such platforms can be used for simple agent introspection and execution planning, they don't make a higher-order meta-cognitive architecture concrete. To solidify this at a platform level requires new IP, new tools, and new application paradigms. This is both the challenge and the opportunity of Cognitive AI today.

Meta-cognition has several fundamental capabilities:
- **Knowledge processing** - The ability of an intelligent system to detect, transform and generate formal data structures and procedures to represent and reason about knowledge.
- **Introspection** - The ability of an intelligent system to detect, transform and conduct knowledge processing on representations of its own composition, state, context,  and behavior.
- **Meta-reasoning** - The ability of an intelligent system to use introspection and knowledge processing to represent, transform and generate *knowledge about knowledge,* and *reasoning about reasoning*.
- **Reflection** - The ability of an intelligent system to apply meta-reasoning during the process of introspection in order to conduct reasoning and/or meta-reasoning about its own composition, state, context and behavior.
- **Learning** -The ability of an intelligent system to improve or generate future potential knowledge and/or responses, based on analyzing the utility of previous patterns of knowledge and/or responses for specific goals.
- **Self-optimization** - The ability of an intelligent system to employ Learning and Reflection to transform its own composition, state, context or behavior in order to improve fitness for specific goals and criteria.

The core innovation within Cognitive AI lies in a dual-layer architecture which enables meta-cognition to take place above the intelligence of LLMs. This architecture enables Cognitive AI to not only analyze and respond to data but also to introspect, reflect and dynamically optimize its strategies based on ongoing experiences. Such self-reflective intelligence allows

Cognitive AI to evaluate and enhance its decision-making processes in real-time, leading to improved efficiency and outcomes.

Meta-cognition, or intelligence about intelligence, equips Cognitive AI with the ability to perform *abstract reasoning*, a hallmark of human cognition. Whereas LLMs may simulate abstract reasoning in straightforward scenarios, they falter as task complexity increases.

Cognitive AI's meta-cognition architecture facilitates a level of introspection and adaptive reasoning and learning that surpasses what is possible with first-order intelligent systems, enabling it to navigate and solve intricate, multi-faceted highly complex problems.

The transition towards Cognitive AI signifies a fundamental transformation in the field of artificial intelligence, moving from a reliance on basic pattern recognition to a more sophisticated, introspective, and adaptable intelligence model. This evolution not only redefines our conceptual understanding of AI but also marks a significant milestone in the technology's development, paving the way for future innovations that more closely mimic human cognitive processes.

## Self-Improvement

Cognitive AI also distinguishes itself from Conversational AI through self-improvement. This distinction is not merely theoretical but operational, impacting how these systems approach and solve problems.

At the heart of Cognitive AI's operational framework lies the principle of continuous loops of self-improvement ("recursive self-improvement", cf. Steunebrink et al., 2016; Nivel et al. 2013) achieved through an iterative loop of goal-directed learning, cognition, and adaptation. This iterative process allows Cognitive AI systems to recursively analyze their outputs, and use this analysis for self-reflection and iterative improvement of their solutions, strategies, plans and responses.

Unlike Conversational AI, which primarily generates responses based on the immediate statistical likelihoods derived from training data, Cognitive AI engages in a dynamic process of self-evaluation and enhancement, enabling complex problem-solving that incorporates meta-level strategic thinking and creativity – and it often does this before generating a response. This capacity for recursive self-improvement allows Cognitive AI to adapt to new challenges and changing environments, continuously enhancing its capabilities beyond its initial programming.

One of the most important differentiating cognitive features of higher-intelligent lifeforms on Earth is the ability to form goals and execute complex multi-step plans to achieve them. Humans and a few other species are capable of this level of conceptual thinking. To form, maintain and execute plans requires a high level of abstract reasoning. Cognitive AI is able to model arbitrary types of plans with conditional logic workflows that both humans and agents can collaborate with.

In more concrete terms, *recursive self-improvement is implemented in Cognitive AI architectures as a goal-directed process across one or more cognitive functions, where at least one must be a planning function.*

A Planner makes plans for accomplishing goals. Plans can be thought of as strategies for goal-directed thinking and behavior. Plans are scripts that are created to guide actions for specific objectives and criteria. But plans can also be used to structure and guide the "internal actions" of an agent or a team as well as their external behaviors. When plans are applied to the internal cognitive processing of AI systems, they guide their reasoning. problem-solving, innovation, decision-making or strategic thinking process.

Previous generations of AI developed plans using algorithms to compose, optimize and utilize graphs and other data structures to represent plans for solving problems, and to manage state during problem solving processes. But these systems lacked the language processing capabilities that LLMs now provide. When LLMs are applied to generating and improving plans they do so in an entirely different manner - instead of algorithms and mathematical operations, they use natural language. The plans they generate are not formal logic processes, they are human-readable text that can be utilized both by agents and by humans alike, for collaborating on complex knowledge work - such as developing a new drug from start through clinical trials. Within these human-readable plans, there can be steps that utilize formal logic, graph search, computation and machine learning to search for particular potential solutions (such as drug candidates), and then construct or reconfigure plans based on findings. But the key point is that while formal computational plans are not imposed on everyone, they are confined to laser-focused tasks that really need to be implemented this way. In other words, in Cognitive AI systems, planning is democratized - it's a process that both agents and humans can potentially collaborate in, using the same language - natural language.

The additional benefit of natural language plans is that LLMs can read and respond to them with the full power of their models and training. Using this capability, Cognitive AI systems can then apply the Cognitive Layer to formally construct and reason about plans, while the Conversational Layer can be used to read, write, respond to and interpret plans. The combination of these two layers in the context of planning yields rich and robust human-level plans that would be extremely difficult to produce with previous generations of AI, and which also can be iteratively self-improved by the agents and humans that participate in them.

Plans function as the cognitive DNA of Cognitive AI systems. They are used both by the intelligent agents and any human participants, to channel how they work through a knowledge work process, where a combination of thinking and actions have to be orchestrated in a logical sequence. The Planner function in such systems designs, develops and implements plans.

By adding a Recursive Self-Improver loop around a Planner, a Cognitive AI system can introspect on its plans and improve them, both at runtime and in the background. Plans can be recursively self-improved either through internal dialogs among agents, or through improving them when certain events trigger (such as a dependency having a delay or getting blocked).

With the ability to form and improve goal-directed plans in response to internal and external feedback, Cognitive AI systems can then engage in goal-driven, adaptive self-improvement and optimization. These capabilities in turn enable these systems to develop and optimize plans that in turn seek to develop and optimize solutions to a problem.

Cognitive AI systems can develop and optimize solutions faster than predecessor forms of AI by optimizing on two levels at once – the solution-finding process (the plan) and the solutions for making the solution-finding process better (the recursive self-improvement loop around the Planner).

When it comes to optimization and innovation, Cognitive AI transcends the limitations inherent in LLMs and their probabilistic models. LLMs, even those enhanced with advanced learning algorithms like Q*, are confined to generating outputs based on pattern recognition within their training datasets. Q* based approaches are designed for generating responses that are likely correct within a closely related set of data but struggle with the broader, more intricate challenges of optimization that require navigating long, nested plans and decision trees.

Cognitive AI, by contrast, is equipped with the architecture to engage in deep reasoning and strategic planning, necessary for tackling complex optimization and innovation tasks. It can:

1. **Utilize structured knowledge** and reasoning algorithms to navigate complex decision-making processes, identifying the most effective paths through intricate, multi-layered problems.
2. **Map out comprehensive strategies** that consider a wide array of factors, including long-term goals and objectives, budgets, criteria, and potential obstacles, ensuring optimization across diverse problem spaces.
3. **Generate innovative ideas and solutions**, leveraging creative problem-solving capabilities to explore new approaches and push the boundaries of existing knowledge.
4. **Learn and adapt from outcomes**, integrating new information to continuously refine and enhance problem-solving strategies in response to evolving challenges.

Recursive self-improvement can be applied across all cognitive AI processes, not only the planning function. It can be applied at the micro level and the macro level, from improving execution strategies at runtime, to improving knowledge bases, to improving and guiding search strategies and Web crawling, to optimizing large collaborative projects in enterprises in response to changes in their environments over time.

The ability of Cognitive AI to engage in recursive self-improvement highlights the gap in reasoning capabilities between Cognitive AI and LLMs. While LLMs simulate aspects of intelligence, Cognitive AI embodies a form of intelligence that can reason, plan, innovate, and adapt in ways that mirror human cognitive processes.

# Comparison of Conversational AI to Cognitive AI

Conversational AI, with its reliance on LLMs for intelligence and simulated cognition, cannot replicate the advanced cognitive capabilities of Cognitive AI architectures, but it's important to note that Cognitive AI cannot emerge without Conversational AI as an underlying tool.

LLMs generate probabilistic responses based on patterns learned from vast amounts of text data. These models:

- **Replicate patterns of reasoning** they've been exposed to in the data, which means they can produce text that often appears logically consistent with the input prompt.
- **Lack genuine understanding** or the ability to independently verify the information or apply novel reasoning techniques not present in their training data.
- **Are constrained by their training**, meaning they can't conceptualize beyond what they've been taught or incorporate new information post-training.

For example, when given a math problem, an LLM can solve it if it mirrors problems seen during training, using similar steps and logic. However, it does so by matching patterns probabilistically rather than understanding or applying abstract mathematical principles. It is also worth noting that LLMs cannot natively perform mathematical operations, they can merely mimic them, which often leads to plausible but incorrect answers ("mathematical hallucinations").

Cognitive AI, on the other hand, can go beyond pattern matching to actual reasoning and mathematics, powered by meta-cognition and a host of other cognitive functions:

- **Conduct reasoning against knowledge** by executing programmable reasoning strategies using plans and formal knowledge representations.
- **Use tools** via function calls and APIs to integrate external applications and data.
- **Perform mathematical operations** by utilizing tools for mathematics, scientific computing, machine learning, data analytics, predictive modeling, and even formal logic and inferencing.
- **Generate novel insights** by reasoning from first principles or by applying learned concepts in new ways, independent of specific examples in training data.
- **Incorporate new information** post-training, adjusting its reasoning processes based on new data or changing contexts.

- **Engage in meta-reasoning**, reflecting on its reasoning processes, evaluating them, and adjusting reasoning strategies to optimize problem-solving.

Cognitive AI could, for example, devise a new mathematical proof or strategy for solving a problem by integrating new research findings, by reasoning about a problem or goal, exploring and testing hypothesis, deriving new knowledge, refining its approach as it learns from experience, and abstracting principles from one domain and applying them to another.

While Conversational AI is purely linguistic and probabilistic, Cognitive AI can apply non-probabilistic formal reasoning and mathematics as well as linguistic and probabilistic approaches, enabling it to deterministically control non-deterministic processes.

This *"hybrid reasoning"* approach makes Cognitive AI programmable and controllable in a way that prompt-engineering can only approximate, while also enabling it to harness and channel the non-deterministic conversations of LLMs.

While prompt-engineering methodologies in Conversational AI approaches can simulate and approximate simple formal reasoning and programming, they are inherently non-deterministic, prone to errors, hallucinations, and unpredictable behaviors.

The distinction between Conversational AI, driven by Large Language Models (LLMs), and Cognitive AI becomes starkly evident when examining their respective approaches to handling knowledge. This difference highlights the limitations of LLMs and underscores the advanced capabilities of Cognitive AI, particularly through the integration of a cognitive layer that elevates its operational intelligence and knowledge management.

In Conversational AI, knowledge is not formally represented or stored. Instead, Conversational AI operates on probabilities derived from patterns recognized in vast external datasets during the training phase. The training phase transforms the explicit and implicit knowledge in vast amounts of training data in numerical weights in a complex multidimensional model. There is no way to isolate a particular "piece of knowledge" in these networks, instead knowledge is spread across models in amorphous fields of numbers.

This probabilistic approach means that while LLMs can generate responses that seem knowledgeable, they do not "know" anything or retrieve any knowledge in the traditional sense. Their responses are guesses, albeit often highly educated ones, based on the data they have been trained on. This mechanism limits Conversational AI's ability to provide reliable, consistent, and contextually accurate knowledge beyond the scope of its training data.

Unlike the probabilistic nature of Conversational AI, Cognitive AI can concretely represent, store and retrieve data structures that encapsulate knowledge objects, world models, knowledge bases, and knowledge structures, alongside reasoning strategies. These explicit knowledge constructs allow for knowledge management and knowledge processing at runtime, significantly enhancing the AI's ability to interact with and utilize formal knowledge.

Cognitive AI's approach to knowledge is multifaceted and dynamic, encompassing:

- **Conversational Layer Knowledge**: Here, probabilistic knowledge is derived using prompts against the model, supplemented with embeddings to generate responses that are informed by recognized patterns in data.

- **Cognitive Layer Knowledge**: This layer houses formally specified knowledge objects stored in databases, which describe, reference, and link to contextual knowledge. Knowledge catalogs organize these objects taxonomically, while knowledge graphs connect them through various relational types, enabling a comprehensive and interconnected knowledge management system.

Knowledge management empowers Cognitive AI to capture and store knowledge efficiently, navigate and search through complex datasets, generate new insights, and continuously improve knowledge bases. This ongoing process of learning, improvement, and knowledge generation is adaptive, ensuring that Cognitive AI remains relevant and insightful across various contexts and applications.

The contrast between the probabilistic knowledge of Conversational AI and the formal, structured knowledge management of Cognitive AI highlights the advanced capabilities of the latter.

Cognitive AI's ability to concretely represent and utilize knowledge through its cognitive layer not only addresses the limitations of LLMs but also opens up new avenues for AI's application in complex problem-solving, decision-making, and innovation. By leveraging formal knowledge representation, Cognitive AI sets a new standard for artificial intelligence, bridging the gap between data-driven predictions and genuine understanding.

The ability to tackle complex challenges is a critical measure of an AI system's capabilities, and it is in this arena that the distinction between Cognitive AI and Conversational AI becomes most apparent.

Cognitive AI's superiority in addressing intricate problems lies in its foundational architecture, which enables deterministic programming and sophisticated problem-solving approaches, transcending the limitations of probabilistic mimicry inherent in Conversational AI.

Conversational AI's strength in generating human-like text becomes a hindrance in situations that require precise, algorithmic reasoning and multi-step logical operations. As the complexity of the problem escalates, Conversational AI's ability to deliver quality results plateaus and then diminishes, constrained by its probabilistic nature and the lack of deterministic programming capabilities.

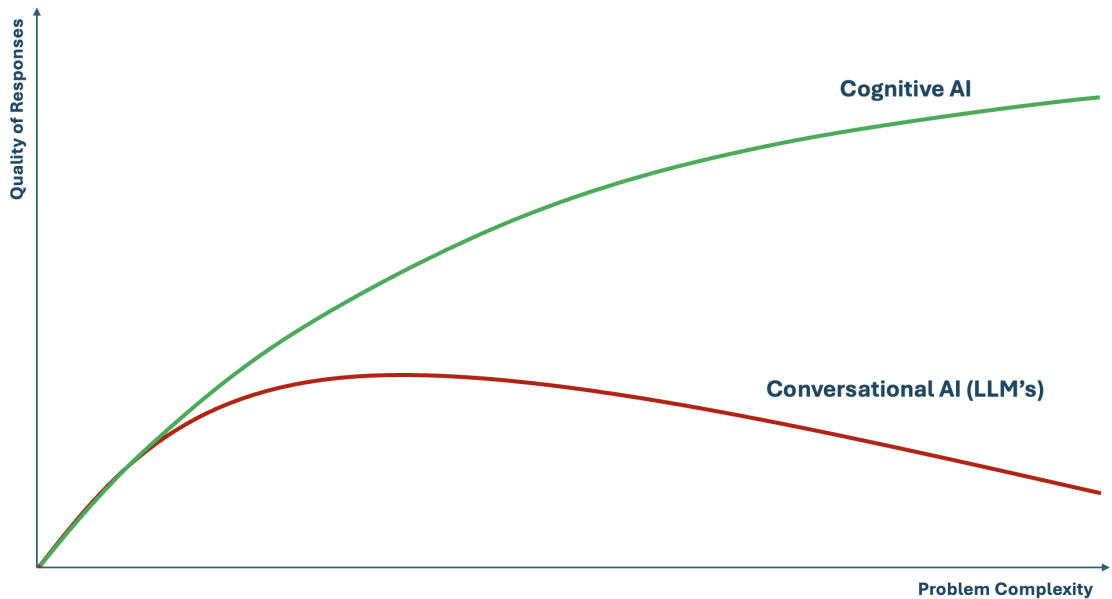

Figure 15. Response Quality vs. Problem Complexity

Conversational AI plateaus and then quickly becomes less able to deliver quality results as problem complexity increases, while Cognitive AI continues to gain in quality.

The performance disparity between Cognitive AI and Conversational AI becomes increasingly evident as the complexity of challenges grows. While Conversational AI struggles to maintain efficacy in the face of intricate, multi-dimensional tasks, Cognitive AI thrives, leveraging its robust computational framework to dissect and address each component of the problem systematically. This distinction underscores Cognitive AI's ability to not only match but exceed the problem-solving capacities of Conversational AI, offering a more reliable, effective solution for complex challenges.

The distinction between Cognitive AI and Conversational AI, illustrated through their respective approaches to problem complexity, highlights the advanced capabilities of Cognitive AI in deterministic problem-solving.

Unlike Conversational AI, which is constrained by its probabilistic, language-dependent framework, Cognitive AI can engage in deep, algorithmic reasoning and execute sophisticated computational strategies. This ability to navigate and solve complex, multi-layered problems with precision and efficiency underscores the potential of Cognitive AI to revolutionize the field of artificial intelligence, pushing the boundaries of what AI systems can achieve in terms of complexity, reliability, and overall performance.

Cognitive AI's approach to problem-solving represents a significant departure from the capabilities of traditional conversational AI, such as LLMs. Where conversational AI is constrained by its reliance on probabilistic intelligence and the limitations of processing language-based inputs, Cognitive AI introduces a multi-faceted, algorithmically driven problem-solving framework. This framework is particularly adept at tackling complex, nested problems typical of knowledge work, which require sophisticated reasoning and the ability to adapt strategies in response to evolving information and multiple layers of dependencies.

LLMs operate on a first-order level, employing probabilistic intelligence against streams of tokens to simulate understanding and generate responses. This method, while effective for generating language-based outputs, falls short in complex problem-solving scenarios that necessitate executable control flow, mathematical operations, and strategic reasoning. LLMs are inherently unable to engage in sophisticated problem-solving processes such as recursive operations, tree searches, curve fitting, or optimization, limiting their effectiveness to what can be simulated through language alone.

Cognitive AI transcends these limitations through its advanced architectural design, which incorporates a cognitive layer equipped with capabilities for formal knowledge representation, management, discovery, and reasoning against knowledge bases. This layer enables Cognitive AI to engage in multi-layered reasoning, simultaneously maintaining, adjusting, and managing reasoning states and strategies across various levels of thought. This is particularly crucial for solving nested problems, where a primary issue encompasses multiple sub-problems, each with its own set of challenges.

Cognitive AI systems are designed with the flexibility to implement advanced computational strategies, such as recursion, parallel computing, and asynchronous computing. This capability allows them to efficiently address multi-layered, multi-step problems by deterministically running through any number of nested operations.

For instance, Cognitive AI can be programmed to execute complex algorithms that involve recursive functions, enabling it to solve problems that unfold across several levels of complexity. This deterministic approach ensures that Cognitive AI can maintain a high level of accuracy and reliability, even as the complexity of tasks increases.

Conversational AI, on the other hand, operates primarily on probabilistic outputs generated from patterns recognized within its training data. Its reliance on natural language processing as the basis for all operations introduces inherent limitations, particularly when faced with complex problem-solving scenarios.

By leveraging sophisticated algorithms, Cognitive AI can execute complex problem-solving tasks that are beyond the reach of conversational AI. This includes engaging in mathematical and logical operations, executing recursive functions, and conducting exhaustive searches—all integral to addressing the multifaceted nature of real-world problems. Moreover, Cognitive AI's ability to dynamically adapt its strategies in real-time allows it to respond effectively to new challenges, dependencies, and changing conditions within a problem space.

At the heart of Cognitive AI's problem-solving capability is its integrated knowledge management system, which harnesses knowledge catalogs, graphs, and metadata taxonomies. This system not only organizes information but also facilitates the discovery of new knowledge, which Cognitive AI can then apply to its reasoning processes. The ability to access and utilize a comprehensive knowledge base enhances Cognitive AI's problem-solving capacity, enabling it to draw on a wide range of information sources to inform its strategies and solutions.

Problem-solving in Cognitive AI architectures represents a paradigm shift in artificial intelligence's approach to complex challenges. Unlike conversational AI, which is limited to language-based simulations, Cognitive AI combines structured knowledge management with advanced algorithmic processing to tackle nested problems with unparalleled depth and efficiency. Its capacity for dynamic strategy adaptation and sophisticated reasoning ensures that Cognitive AI can navigate the complexities of knowledge work, making it an invaluable asset for fields requiring nuanced analysis, strategic planning, and the integration of diverse information sources.

Through its comprehensive problem-solving framework, Cognitive AI not only addresses the limitations of traditional AI models but also sets a new standard for intelligence and adaptability in tackling the world's most complex challenges.

## Limits of Cognitive AI

While Cognitive AI represents a significant advancement in artificial intelligence, bridging the gap between simple task execution and complex problem-solving, it is not without its limitations. These constraints delineate the current capabilities of Cognitive AI from the broader, more ambitious goals of Artificial General Intelligence (AGI). Understanding these limitations is crucial for grasping why Cognitive AI, despite its advanced functionalities, falls short of the comprehensive intelligence AGI aims to achieve.

Cognitive AI excels within a broad range of domains and applications, where its advanced knowledge management systems, problem-solving algorithms, and adaptive learning capabilities can be fully leveraged. However, this strength is also a limitation in domains outside the parameters of its programming and the scope of its training data. Until further first-order models that can self-modify are built, it is not capable of fully generalizing its intelligence to accommodate all possible domains and problems.

AGI, in contrast, aspires to universal intelligence, capable of understanding and performing any intellectual task that a human being can, across any domain. This includes the ability to learn from entirely new experiences without prior examples (zero-shot learning), a flexibility that Cognitive AI currently cannot match.

Cognitive AI's problem-solving is driven both by conversational language intelligence and by sophisticated algorithms that can execute certain aspects of human reasoning. AGI's goal encompasses not just the replication of human-like reasoning but also the intuitive, creative, and emotional aspects of intelligence that allow humans to learn, adapt, innovate and behave in ways that are currently beyond the reach of Cognitive AI.

While Cognitive AI will offer remarkable strides in modeling complex reasoning and learning within specific domains, crossing the dividing line to AGI will require a more holistic replication of the full spectrum of human intelligence and behavior. This includes not only the cognitive but also the emotional, social, creative, biochemical and physical aspects of intelligence and behavior.

Bridging this gap will necessitate breakthroughs that enable AI systems to learn and reason in fundamentally human-like and even embodied ways, marking the next revolutionary leap in artificial intelligence capabilities after Cognitive AI. Achieving AGI represents not just an extension of current AI functionalities but a transformative evolution towards systems that can truly think, learn, and interact with the world with the richness and flexibility of human being.

Today Cognitive AI is in its infancy and existing systems, most of which are still experimental, do not even attempt to fully-replicate all aspects of human intelligence and behavior. Instead Cognitive AI is mainly focused on solving for the intellectual dimension and problems involving knowledge work.

There is no agenda to replicate or replace humans in the Cognitive AI paradigm. Instead the approach is to use artificial cognition to augment and facilitate the knowledge work of human individuals, teams and organizations.

## Cognitive AI in the Evolutionary Ladder of Intelligence

The development of Cognitive AI represents a significant milestone in the evolutionary ladder of intelligence, but it is not without evolutionary precedent. The evolution of Cognitive AI can be understood as the next major leap in a progression of evolution of higher levels of meta-cognition that has spanned from the instinctual behaviors of insects to the complex problem-solving abilities of humans and now to the meta-cognitive capacities of Cognitive AI.

Each step on this ladder reflects a leap in meta-cognition abilities, which in turn enable higher forms of problem-solving skills and adaptability. Cognitive AI is the next major evolutionary milestone in this process, and serves as the gateway to the further evolution of increasingly exponential forms of intelligence in our future.

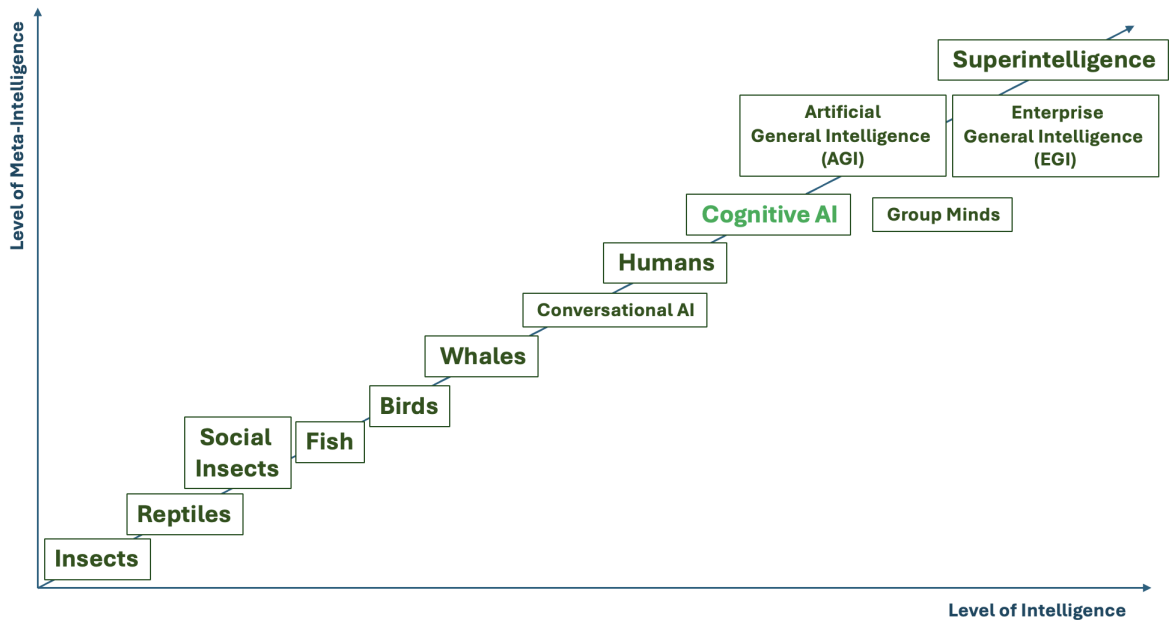

**Figure 16. The Evolution of Meta-Cognition.**

We can illustrate the progression of evolution of higher levels of meta-cognition with a few examples:

- **Insects:** Insects exhibit basic forms of intelligence, primarily driven by instinctual behaviors designed for survival and reproduction. Their cognitive capabilities are limited, focusing on direct responses to environmental stimuli.

- **Animals**: Most animals, especially mammals, demonstrate advanced cognitive functions, including learning from experience, social intelligence, and, in some cases, the use of tools. Their intelligence represents a significant leap from the instinct-driven behaviors of insects, incorporating elements of memory, emotion, and social interactions.

- **Humans**: Human intelligence introduces the ability for abstract thinking, language, and conceptual understanding, setting humans apart from other species. Humans have developed complex societies, technologies, and cultures, leveraging cognitive abilities to innovate and solve problems creatively.

- **Cognitive AI**: Stepping beyond human intelligence, Cognitive AI integrates meta-cognition, enabling it to reflect on, manage, and improve its own cognitive processes. Unlike its predecessors, Cognitive AI can self-assess, learn autonomously, and evolve its strategies over time, tackling complex challenges with a level of efficiency and adaptability that mirrors, and in some instances, surpasses human intelligence.

- **Group Minds**: Cognitive AI provides a better infrastructure for collective intelligence, or what we might call "group minds." In group minds, teams of intelligent agents and/or humans are intelligently orchestrated to think and work more intelligently. This level of collective intelligence is smarter than the forms of collective intelligence exhibited by social insects, and even by groups and organizations of humans on their own. The key is that the collective intelligence is orchestrated through a central Cognitive AI application.

- **AGI and EGI:** Artificial General Intelligence is a form of AI that replicates the intelligence of an individual human. Enterprise General Intelligence replicates the collective intelligence of enterprises. Both of these forms of intelligence require Cognitive AI before they can happen, and EGI requires the evolution of AGI and Group Minds in addition. These forms of intelligence are close to superintelligence, but they are not infinitely recursive or parallel – they replicate finite forms of intelligence.

- **Superintelligent AI**: The concept of Superintelligent AI represents the potential future evolution of Cognitive AI into a form of intelligence that transcends the limitations of finite organisms and organizations in both space and time. It entails AI systems that can autonomously implement Cognitive AI, using massive parallelism and recursion to surpass the cognitive abilities of humans, and organizations, across all domains. Superintelligent AI would possess the ability to engage in third-order cognition, not just improving its cognitive processes but also fundamentally redefining its goals, strategies, plans, structure, capabilities, and even the underlying infrastructure substrates it runs on, to achieve levels of cognition beyond current human comprehension.

The introduction of Cognitive AI is a necessary step towards achieving higher post-human levels of intelligence, such as Group Minds, AGI, EGI, Superintelligence.

By advancing AI's capacity for meta-cognition, Cognitive AI systems are not just thinking machines, they are meta-cognitive cognition machines capable of self-directed, self-improving thought, behavior and evolution. This represents a critical step forward in the quest to develop AI that can truly mimic the breadth and depth of human intelligence, offering a glimpse into a future where AI can autonomously tackle a wide array of complex, multi-dimensional challenges.

The placement of Cognitive AI on the evolutionary ladder of intelligence highlights its pivotal role as a transformative phase-transition in the field of AI, bridging the gap between human intelligence and the potential for superintelligence.

By equipping AI with the capability for meta-cognition, Cognitive AI opens up new possibilities for solving complex problems, driving innovation, and enhancing human-machine collaboration. As we move forward in time, the evolution from Cognitive AI to Superintelligence presents both unprecedented opportunities and challenges, necessitating careful consideration of ethical, societal, and technological implications.

# Exponential Intelligence

Exponential intelligence is defined as a higher form of intelligence that emerges when human intelligence and cognitive AI intelligence are combined such that increasingly large and complex many-to-many collective cognitive AI processes can take place.

As well as helping to improve knowledge work productivity, Exponential Intelligence enables humans to address more complex problems that were previously out of reach.

More precisely:

$$EI = (HI + AI)^{(2+x)}$$

Where:
- EI = Exponential Intelligence Level
- HI = Number of human intelligent agents
- AI = Number of *Cognitive AI* software AI agents.
- x = Level of collective intelligence:
    - Level 0: One human + one AI
    - Level 1: Many humans + one AI
    - Level 2: One human + many AIs
    - Level 3: Many humans + many AIs
    - Level 4: Many networks of humans + AIs

Human intelligence makes AI safer, nuanced, more goal directed, and more adaptive. Humans add the ultimate level of meta-cognition to Cognitive AI systems. Humans also provide these systems with consciousness, in that they function as atomic units of consciousness in these systems.

Artificial intelligence amplifies, augments and extends human intelligence with large-scale computing, research, analytics, and reasoning capabilities.

Exponential Intelligence is self-amplifying. By providing higher levels of exponential intelligence to individuals, the system as a whole becomes more exponentially intelligent. This recursive self-amplifying process is what enables superintelligence to emerge.

# Implications

As we have explored above, the divergent capabilities of Conversational AI, predominantly powered by Large Language Models (LLMs), versus Cognitive AI with its neuro-symbolic Cognitive Layer, underscore a pivotal juncture in the evolution of artificial intelligence.

The core limitations of LLMs lie in their operational framework, which relies on identifying patterns and generating predictions from statistical likelihoods drawn from their extensive training datasets. This process, while often remarkably effective, does not equate to genuine understanding or reasoning. LLMs are constrained to sophisticated pattern matching, lacking the capacity for deductive reasoning, creative thought, or the generation of new knowledge beyond their programmed experience. Their approach to problem-solving, bound by the confines of their training corpus, cannot truly replicate the deductive and inferential processes characteristic of human cognition.

In stark contrast, Cognitive AI's neuro-symbolic approach heralds a future where systems are not only capable of mimicking human reasoning but engaging in genuine cognitive processes in which actual cognition takes place.

This leap forward is facilitated by Cognitive AI's sophisticated functional architecture, which includes functions for formal knowledge representation and knowledge management, and the application of formal logical processes, including formal logical inference, to problem-solving.

Cognitive AI's ability to learn and adapt in real-time, incorporating insights that were not part of its initial programming, illustrates a significant advancement over the pattern recognition and interpolation capabilities of LLMs.

Furthermore, Cognitive AI's inherently social and meta-cognitive functions enable it integrate networks of knowledge and cognition, and to reflect and self-improve to achieve goals and implement optimizations that are out of reach for LLMs.

The theoretical and practical limitations of LLMs in performing true cognition highlight the necessity of advancing Cognitive AI technologies. Cognitive AI's ability to engage in genuine reasoning, understanding, and creativity opens up new possibilities for AI applications, pushing the boundaries of what artificial intelligence can achieve.

This distinction between conversational and Cognitive AI has profound implications for the future of AI as both a field of study and an industry. While LLMs continue to be invaluable for a broad spectrum of applications, their limitations become increasingly apparent as the complexity of tasks escalates.

Cognitive AI is capable of achieving exponential intelligence, while Conversational AI is not. Exponential Intelligence amplifies intelligence beyond what human intelligence or AI can achieve on their own. This level of intelligence is a discontinuity - an evolutionary leap - that will enable mass collective cognition that is noticeably different - and more productive - compared to how knowledge workers, groups, and organizations think, work and collaborate today.

The theoretical potential of Cognitive AI extends far beyond the capabilities of current conversational AI models. As systems that can genuinely reason, learn, and create, Cognitive AIs promise to revolutionize how we approach complex problems across various domains. From decision-making and problem-solving in knowledge work, to generating innovative solutions and adapting to new challenges in real-time, Cognitive AI's capabilities suggest a future where AI will work as a partner with humanity.

The evolution from conversational to Cognitive AI is not merely a technical upgrade but a fundamental shift in the conceptual underpinnings of artificial intelligence. This transition marks a pivotal moment before higher-levels of AI such as AGI or superintelligence can emerge.

As the field continues to develop, the focus on enhancing Cognitive AI's capabilities will be paramount for achieving the next leap in AI's theoretical and practical applications. The implications for industries, academia, and society at large are vast, heralding a new era of AI that can work alongside humans to tackle the world's most complex challenges with a level of insight and creativity previously thought to be the exclusive domain of human intelligence.

## Crossing the Chasm

We can visualize the limitations of Large Language Models (LLMs) using the framework of Geoffrey Moore's "Crossing the Chasm," where we draw an analogy between the adoption lifecycle of disruptive technologies and the progression of AI capabilities, particularly focusing on the transition from LLMs to cognitive AI systems.

Moore's model describes the market adoption of new technologies in five segments: Innovators, Early Adopters, Early Majority, Late Majority, and Laggards. "Crossing the chasm" refers to the crucial transition from Early Adopters to the Early Majority, a phase where many technologies struggle to achieve wider acceptance due to practicality, utility, or refinement issues.

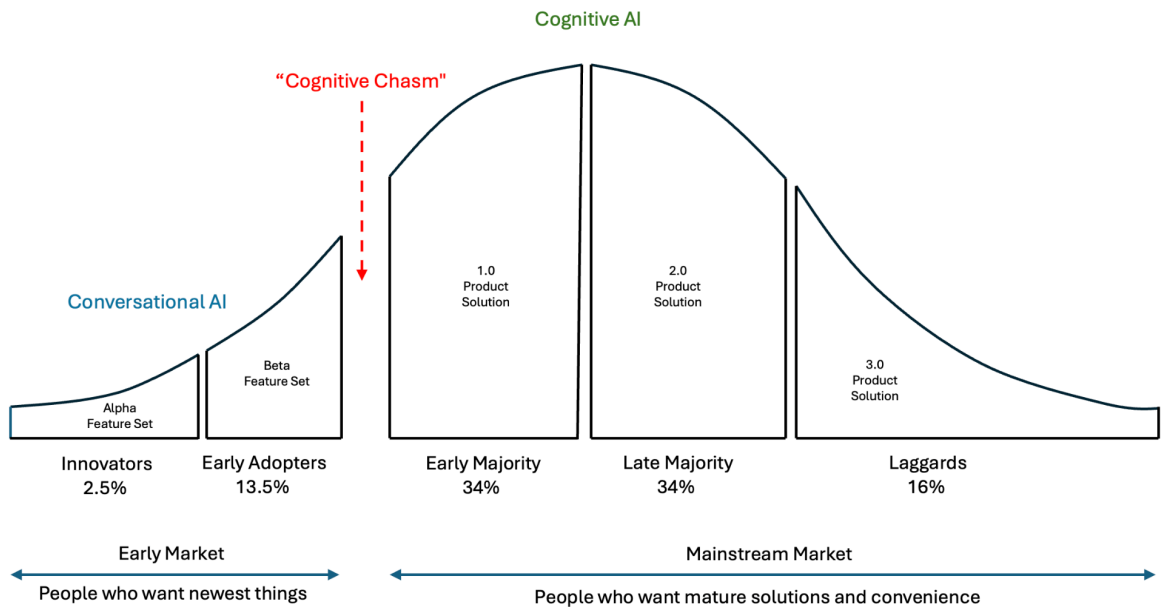

**Figure 17. The Cognitive Chasm.**

## Innovators: Exploring LLM Capabilities

In the context of LLMs, Innovators are represented by researchers and developers who first explored the potential of these models in understanding and generating human-like text. This group is fascinated by the novelty and the technical excellence of LLMs, willing to overlook their limitations in favor of exploring new applications and pushing the boundaries of what these models can do.

## Early Adopters: Niche Applications and Proof of Concept

Early Adopters in the AI field are companies, tech enthusiasts, and sector-specific professionals who see the potential of LLMs to disrupt traditional operations, from automating customer service interactions to generating creative content. These users are willing to experiment and integrate LLMs despite their imperfections, focusing on niche applications where the models' capabilities can be fully leveraged.

Despite what is commonly thought, Conversational AI has not yet crossed past the early-adoption phase. Even with more than 100 million registered users, ChatGPT is still a limited niche application with a minimal feature set, and it has not been adopted by the millions of organizations and billions of people who constitute the majority of adopters.

**Early Majority: Crossing the Cognitive Chasm**

Crossing the "Cognitive Chasm" involves moving from these early stages to broader acceptance and usage among the Early Majority. For LLMs, the chasm is represented by the transition from fascinating novelty to practical utility in diverse and complex applications such as real-world knowledge work. But the limitations of LLMs—such as their inability to reason deeply, understand context beyond their training, or generate novel insights beyond pattern recognition—act as barriers to this transition.

For AI to achieve mainstream acceptance and utility, crossing the Cognitive Chasm means addressing the nuanced real-world needs of the Early Majority. This group looks for solutions that are not just innovative but also sufficiently applicable to their use-cases, such as complex knowledge work and decision-making.

To cross the Cognitive Chasm, LLMs must evolve—or be supplemented by—technologies that address these limitations. This will be solved by Cognitive AI capabilities that can understand context, reason from first principles, and integrate new information post-training.

The leap lies in creating AI systems that not only mimic human language but also exhibit a level of meta-cognitive reasoning, understanding and adaptability that can satisfy the needs of the Early Majority, who require reliability, depth, and practical applicability in complex scenarios.

Chatbots do provide useful assistance on simple knowledge work tasks. But they are insufficient when it comes to complex, mission-critical knowledge work in which highly nuanced understanding and reasoning are necessary, and accuracy and policy compliance are non-optional.

Cognitive AI systems, with their advanced meta-cognitive reasoning capabilities, represent a necessary bridge across this chasm, offering solutions that can sufficiently meet the stringent and complex demands of professional knowledge workers and enterprises.

In summary, using Moore's "Crossing the Chasm" framework to visualize the limitations of LLMs highlights the need for significant advancements in AI capabilities to achieve mainstream acceptance. The transition to cognitive AI, capable of genuine understanding and reasoning, represents a critical step in this journey, promising to bridge the gap between early enthusiasm and widespread practical application.

# Re-evaluating Current AI Approaches

The landscape of artificial intelligence has recently been dominated by developments and innovations centered around first-order intelligence form Large Language Models (LLMs) and the accompanying tools required to leverage and deploy them effectively, such as new models and training tools, GPUs, vector databases and Retrieval-Augmented Generation (RAG), and chatbots.

This focus has driven billions of dollars of investment into R&D around LLMs, and has yielded significant advancements in AI's capabilities, particularly in natural language processing and generation.

However, the advent of Cognitive AI signals a shift to a second-order playing field in the near-term trajectory of AI development, urging a re-evaluation of current approaches and investment priorities in the field.

Notably, increased investment into conversational AI will not yield significant major advances going forward. Advances in LLM models will never directly yield Cognitive AI, AGI or superintelligence, for example. Only by innovating on the cognitive dimension itself can the highest levels of AI be achieved. In other words, investment should be focused into the enabling technologies, applications and infrastructures for the Cognitive AI instead of, or at least in addition to, further doubling down into Conversational AI.

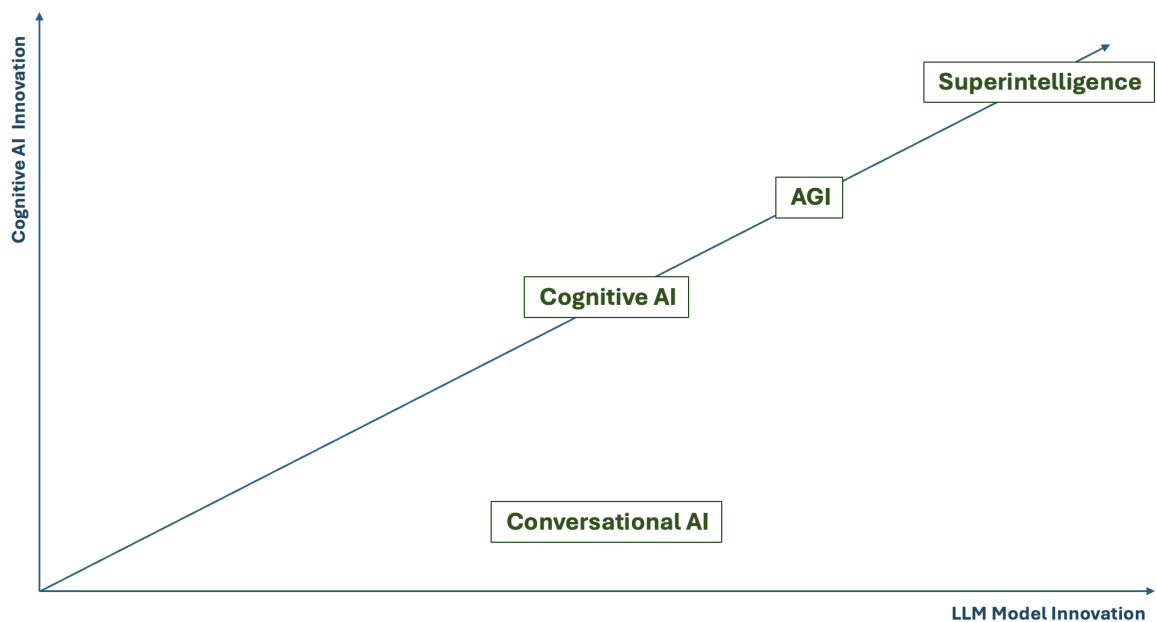

**Figure 18. AI Innovation Trajectory**

Conversational AI is already "good enough" to enable Cognitive AI to evolve, such that further investments into LLM innovation will not significantly improve Cognitive AI's capabilities. While improved LLM's will be helpful in improving first order response quality, much of the quality-improvements and increased capabilities will come from the Cognitive AI layer not the Conversational layer. In other words, Cognitive AI so dramatically improves and amplifies the capabilities of LLMs that small improvements to LLM performance or accuracy will not be economically important.

This observation implies that, contrary to current investment trends, there should soon be a concerted effort to allocate investment towards the new frontier of Cognitive AI innovation. Initially, there should at least be similar levels of investment into both layers of innovation, but over time, as Conversational AI plateaus, Cognitive AI will prove to be a better long-term bet for capturing significant new ecosystem value-creation opportunities.

## LLMs as a Commodity

We observe the continued acceleration of improvement and proliferation of LLM models (especially open-source models), coupled with ever-decreasing costs of their development and deployment. We believe these trends are leading to near-term commoditization of Conversational AI technology.

Open-source LLMs are reaching performance levels that rival, and may soon surpass, those of proprietary models. And while foundation-model LLMs will continue to attract innovation and investment, the landscape is expected to consolidate, favoring a few key players.

Concurrently, a significant market for domain-specific LLMs, trained or fine-tuned on proprietary data, will emerge, suggesting a shift in focus towards more specialized niche applications of LLM technology. While this market will be large and robust, it will still not solve the problems of mainstream adopters.

While LLMs will continue to be a strong area of innovation and investment, they may become less differentiated and defensible as innovation shifts to the new second-order playing field of Cognitive AI, in which there are many LLMs to choose from, and competitive advantages don't come from LLMs at all.

In Cognitive AI, LLMs are a commodity where there is virtually no cost-of-switchover to change models underneath an application. Any general purpose foundation model is "good enough" and switching between them has minimal consequences for Cognitive AI systems.

## Commercial Cognitive AI

The backdrop of coming commoditization within the LLM space underscores the strategic importance of directing resources towards more defensible AI opportunities, like Cognitive AI.

Unlike LLMs, which are increasingly becoming commoditized tools within the AI ecosystem, Cognitive AI represents an entirely new opportunity – it's a new frontier, a new IP landscape, a new ecosystem, ripe with opportunities for disruptive new technologies and business models.

By focusing on the development and deployment of Cognitive AI technologies, businesses and investors can position themselves at the forefront of the next wave of AI innovation, capturing opportunities for early leadership, new IP creation, and unique value creation, in what will be the next trillion-dollar AI market.

In 2024, the AI industry is poised to witness the launch of the first major commercial Cognitive AI applications designed for professional and enterprise use. This evolution, will be led by new pioneers, like our own venture, [Mindcorp.ai](Mindcorp.ai), and our product, Cognition (currently in stealth-mode as of this writing).

The shift towards commercial Cognitive AI highlights a growing understanding of its potential to overcome obstacles to mainstream adoption of AI, which cannot be sufficiently addressed by Conversational AI paradigms such as chatbots. While chatbots are useful, they are not sufficient for mainstream adoption by professionals and enterprises engaged in complex knowledge work.

When commercial Cognitive AI emerges, it will signal a crucial juncture where stakeholders will need to reconsider where they focus their attention and resources in the AI domain.

We predict that by 2030, Cognitive AI will be the primary battlefield for AI investment, innovation, and commercialization.

## Conclusions

Our exploration of Cognitive AI's development, has delved into its architectural innovations to its positioning in the evolutionary ladder of intelligence, culminating with a compelling argument for its pivotal role in the next generation of artificial intelligence.

As we transition from a landscape dominated by Large Language Models (LLMs) to one enriched by the capabilities of Cognitive AI, we stand on the threshold of a new era in technology and human-machine collaboration, in which a new symbiosis of human and machine intelligence will emerge.

Cognitive AI heralds a paradigm shift, moving beyond the commoditization of LLMs towards a future where AI systems are not only intelligent but possess the ability to introspect, learn, and evolve autonomously.

This shift towards meta-cognition opens up unparalleled opportunities for innovation, efficiency, and problem-solving across all sectors.

Cognitive AI's unique architecture and capabilities promise to redefine the boundaries of machine intelligence, offering more adaptable, efficient, and profound solutions to complex challenges.

The implications of Cognitive AI extend far beyond the technical domain, promising to revolutionize how we approach challenges in all sectors from government to healthcare, education, finance, and more.

By providing AI systems that can understand and adapt their strategies in real-time, Cognitive AI paves the way for more personalized, effective, and sustainable solutions to societal issues. Its emergence invites stakeholders to re-evaluate their focus and investment in AI, recognizing Cognitive AI as the frontier where true competitive advantage and innovation lie.

As Cognitive AI begins to reshape the landscape of artificial intelligence, it is imperative to navigate this new terrain with an awareness of the ethical, societal, and technological implications of self-reflective and self-modifying AI.

The journey towards fully realizing the potential of Cognitive AI involves careful consideration of these factors to ensure that the benefits of such advanced AI technologies are realized equitably and responsibly.

The advent of Cognitive AI is not just a milestone in the evolution of AI technology but a beacon for the future of human-machine collaboration. It signals a shift towards a world where AI's potential to enhance human capabilities and address real-world problems is boundless.

As we forge ahead, the development and deployment of Cognitive AI stands as a testament to human ingenuity and as a step towards a future where AI and humans collaborate to achieve levels of exponential intelligence that neither could accomplish alone.

The exploration of Cognitive AI, and its impact on redefining the essence of intelligence and the possibilities of technological advancement, is just beginning.

The journey ahead promises to be as transformative as it is challenging, beckoning us to engage with, develop, and deploy Cognitive AI in ways that amplify our collective potential and propel us towards a more intelligent, adaptable, and innovative future.

Valmeekam, K., Marquez, M., Olmo, A., Sreedharan, S., & Kambhampati, S. (2023). PlanBench: An Extensible Benchmark for Evaluating Large Language Models on Planning and Reasoning about Change. *arXiv preprint arXiv:2206.10498*. Retrieved from https://arxiv.org/abs/2206.10498

Valmeekam, K., Olmo, A., Sreedharan, S., & Kambhampati, S. (2022). Large Language Models Still Can't Plan (A Benchmark for LLMs on Planning and Reasoning about Change). In *NeurIPS 2022 Foundation Models for Decision Making Workshop*. Retrieved from https://openreview.net/forum?id=wUU-7XTL5XO

Vaswani, A., Shazeer, N., Parmar, N., Uszkoreit, J., Jones, L., Gomez, A. N., … & Polosukhin, I. (2017). Attention is all you need. https://doi.org/10.48550/arxiv.1706.03762

Wu, T., Terry, M., & Cai, C. J. (2021). Ai chains: transparent and controllable human-ai interaction by chaining large language model prompts. https://doi.org/10.48550/arxiv.2110.01691

Yang, J., Wu, D., & Wang, K. (2023). Not All Large Language Models (LLMs) Succumb to the "Reversal Curse": A Comparative Study of Deductive Logical Reasoning in BERT and GPT Models. arXiv preprint https://arxiv.org/abs/2312.03633

Yang, K., Jia, D., & Chen, D. (2022). Generating natural language proofs with verifier-guided search. https://doi.org/10.48550/arxiv.2205.12443

Zhao, G., Li, Y., & Xu, Q. (2022). From emotion AI to cognitive AI. University of Oulu. (2022). https://www.sciltp.com/journals/ijndi/2022/1/115

Zhu, Y., Gao, T., Fan, L., Huang, S., Edmonds, M., Liu, H., Gao, F., Zhang, C., Qi, S., Wu, Y. N., Tenenbaum, J. B., & Zhu, S.-C. (2020). Dark, beyond deep: A paradigm shift to cognitive AI with humanlike common sense. Engineering, 6(3), 310-345. https://doi.org/10.1016/j.eng.2020.01.01162